\pdfoutput=1

\documentclass[runningheads]{llncs}
 
\usepackage{eccv}

\usepackage{eccvabbrv}

\usepackage{graphicx}
\usepackage{booktabs}

\usepackage[accsupp]{axessibility}  

\usepackage{hyperref}

\usepackage{orcidlink}

\newcommand{\ent}{\mathcal{H}}

\usepackage{pifont}
\newcommand{\cmarkg}{\color{gray}\ding{51}}%
\newcommand{\xmark}{\ding{55}}%

\usepackage{tikz}
\usepackage{pgfplots}
\usetikzlibrary{shapes.geometric,shapes.multipart,arrows,positioning,calc,shapes.arrows,arrows.meta,pgfplots.statistics,backgrounds}
\pgfplotsset{compat=1.17}
\usepackage{makecell}
\usepackage[inline]{enumitem}

\newcommand{\ours}[0]{AdaGlimpse}

\newcommand{\dcam}{d_{\text{cam}}}
\newcommand{\dmax}{d_{\text{max}}}
\newcommand{\dmin}{d_{\text{min}}}
\newcommand{\dpatch}{d_{\text{patch}}}

\newcommand\smallimg{2cm}
\newcommand\smallerimg{1.4cm}
\newcommand\smallestimg{1.1cm}
\newcommand{\tabfigure}[2]{\raisebox{-.5\height}{\includegraphics[#1]{#2}}}

\definecolor{crimson2143940}{RGB}{214,39,40}
\definecolor{darkgray176}{RGB}{176,176,176}
\definecolor{darkorange25512714}{RGB}{255,127,14}
\definecolor{forestgreen4416044}{RGB}{44,160,44}
\definecolor{gray}{RGB}{128,128,128}
\definecolor{mediumpurple148103189}{RGB}{148,103,189}
\definecolor{steelblue31119180}{RGB}{31,119,180}

\usepackage{array}
\newcolumntype{M}[1]{>{\centering\arraybackslash}m{#1}}

\DeclareMathOperator*{\argmax}{arg\,max}

\begin{document}

\title{\ours{}: Active Visual Exploration with Arbitrary Glimpse Position and Scale} 

\titlerunning{\ours{}}

\author{Adam Pardyl\inst{1,2,3}\orcidlink{0000-0002-3406-6732} \and %
Michał Wronka\inst{2}\orcidlink{0009-0002-5208-9900} \and %
Maciej Wołczyk\inst{1}\orcidlink{0000-0002-3933-9971} \and %
Kamil Adamczewski\inst{1}\orcidlink{0000-0002-2917-4392} \and %
Tomasz Trzciński\inst{1,4,5}\orcidlink{0000-0002-1486-8906} \and %
Bartosz Zieliński\inst{1,2}\orcidlink{0000-0002-3063-3621}} %

\authorrunning{A.~Pardyl et al.}

\institute{IDEAS NCBR \\
\email{\{adam.pardyl, maciej.wolczyk, kamil.adamczewski,\\tomasz.trzcinski, bartosz.zielinski\}@ideas-ncbr.pl} \and
Jagiellonian University, Faculty of Mathematics and Computer Science \\
\email{michal.wronka@student.uj.edu.pl}\and
Jagiellonian University, Doctoral School of Exact and Natural Sciences \and
Warsaw University of Technology \and
Tooploox}

\maketitle

\begin{abstract}
Active Visual Exploration (AVE) is a task that involves dynamically selecting observations (glimpses), which is critical to facilitate comprehension and navigation 
 within an environment. While modern AVE methods have demonstrated impressive performance, they are constrained to fixed-scale glimpses from rigid grids. In contrast, existing mobile platforms equipped with optical zoom capabilities can capture glimpses of arbitrary positions and scales.
To address this gap between software and hardware capabilities, we introduce \ours{}. It uses Soft Actor-Critic, a reinforcement learning algorithm tailored for exploration tasks, to select glimpses of arbitrary position and scale. This approach enables our model to rapidly establish a general awareness of the environment before zooming in for detailed analysis.
Experimental results demonstrate that \ours{} surpasses previous methods across various visual tasks while maintaining greater applicability in realistic AVE scenarios.

\end{abstract}

\begin{figure}[th]
    \centering
    \includegraphics[scale=1]{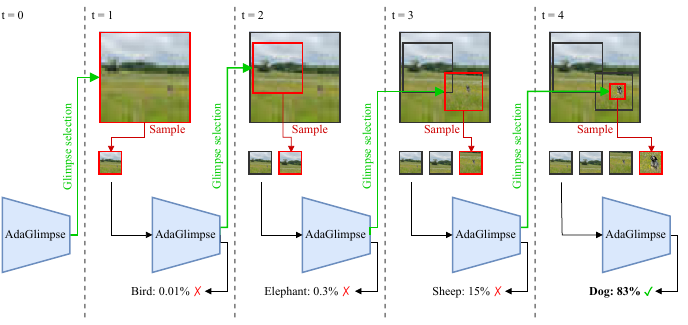}
    \vspace{-.2cm}
    \caption{\textbf{Adaptive Glimpse (\ours{})}: Our approach selects and processes glimpses of arbitrary position and scale, fully exploiting the capabilities of modern hardware. In this example, \ours{} selects a low-resolution glimpse of the whole environment. Based on this glimpse, it predicts a bird with probability $0.01$, too low to make the final decision. Instead, it selects the second glimpse by zooming in to the upper left corner. The process repeats four times until the probability of the predicted class is higher than a specified threshold.}
    \label{fig:teaser}
\end{figure}

\section{Introduction}
\label{intro}

Common machine learning solutions for computer vision tasks, such as classification, segmentation, or scene understanding, usually presume access to complete input data~\cite{girdhar2022omnivore}. However, this assumption does not apply to embodied agents functioning in the real world. Agents such as robots and UAVs face constraints on their data-gathering capabilities, such as a restricted field of view and limited operational time, caused by dynamically changing environments~\cite{4seasons}. Moreover, capturing and analyzing high-resolution images of the entire visible area is inefficient, as not every part of an image contains the same amount of details.

Active Visual Exploration (AVE) addresses the challenge of how an agent should select visual information from its environment to achieve a particular objective. Instead of systematically sampling and analyzing the entire available environment at the highest resolution, an agent dynamically chooses the location for sampling subsequent observations, informed by insights from prior exploration steps~\cite{ramakrishnan2021exploration}.  This process of selecting visual samples, often referred to as \textit{glimpses}, is inspired by the natural way humans explore their surroundings by instinctively moving their heads and eyes~\cite{hayhoe2005eye}.

Current research in active visual exploration can be categorized into two groups. The first group of approaches divides the image into a regular grid of fixed-sized glimpses from which the model tries to pick the most informative ones~\cite{seifi2020attend,seifi2021glimpse,rangrej2022consistency,pardyl2023active}. The second group starts by capturing a low-resolution image of the entire environment, and then it again selects glimpses from regular grids~\cite{elsayed2019saccader,papadopoulos2021hard,uzkent2020learning}.
Relying solely on regular grids fails to fully exploit the capabilities of modern hardware, which can provide a glimpse of any position and scale. For example, in a pan-tilt-zoom camera, we can achieve it by using optical zoom~\cite{DoubleRobotics2024}, while in a UAV, we can alter its altitude~\cite{krotenok2019change}.

In this paper, we overcome current limitations by introducing~\textbf{\ours{}} (Adaptive Glimpse)\footnote{Source code is available at \url{https://github.com/apardyl/AdaGlimpse}}, 
an active visual exploration method that selects glimpses of arbitrary scale and position, significantly reducing the number of observations required to understand the environment. Drawing inspiration from~\cite{pardyl2023beyond}, we build our network on an input-elastic vision transformer. In each exploration step, our model predicts the optimal position and scale for the next glimpse as a value in a continuous space. Since the patch-sampling operation is not differentiable, we train the model using a reinforcement learning algorithm. In particular, we use the Soft Actor-Critic algorithm~\cite{haarnoja2018soft} since it excels in exploration tasks.

Through exhaustive experiments, we show that \ours{}  outperforms state-of-the-art methods on common benchmarks for reconstruction, classification, and segmentation tasks. As such, our method enables more effective utilization of embodied platform capabilities, leading to faster environmental awareness. Our contributions can be summarized as follows:
\begin{itemize}
\item We introduce a novel approach to Active Visual Exploration (AVE) that selects and processes glimpses of arbitrary position and scale.
\item We present a task-agnostic architecture based on a visual transformer.
\item We formulate AVE as a Markov Decision Process with a carefully designed observation space, and leverage the Soft Actor-Critic reinforcement learning algorithm that excels in exploration.
\end{itemize}

\section{Related work}

\subsubsection{Missing data.}
The problem of missing data in context of images has been addressed in a variety of ways, such as inferring remaining information from the input distribution using a fully connected network~\cite{smieja2018processing}, or more commonly by image reconstruction. 
In particular, MAT\cite{li2022mat} performs inpainting using a transformer network with local attention masking, a solution that additionally reduces computation by processing only informative parts of the image. 
A similar principle can be found in ViT based Masked Autoencoder (MAE)\cite{he2022masked} where the encoder network operates only on visible patches, while the decoder processes all patches, including the masked ones.
\subsubsection{Region selection.}
Numerous methods exist for selecting the most informative regions from an image, including expectation maximization\cite{ranzato2014learning,yoo2015attentionnet}, majority voting \cite{alexe2012searching}, wake-sleep algorithm \cite{ba2015learning}, sampling from self-attention or certainty maps\cite{seifi2021glimpse,seifi2020attend,pardyl2023active}, and Bayesian optimal experiment design\cite{rangrej2021probabilistic}; 
yet, recently the most predominant solution is using reinforcement learning algorithms, such as variants of the Policy Search\cite{xu2015show,mnih2014recurrent,mathe2016reinforcement} 
Deep Q-Learning\cite{caicedo2015active,chai2019patchwork} or Actor-Critic\cite{rangrej2022consistency}.  

\subsubsection{Variable scale transformers.}
A number of studies have been conducted to overcome the constraint of Vision Transformers (ViTs) of working only with rigid grid of fixed size patches, be it by modifying grid scale sampling during training phase\cite{li2021efficient,touvron2022deit,wu2020multigrid} or with position and patch encoding rescaling tricks \cite{beyer2022flexivit}. 
Beyond Grids\cite{pardyl2023beyond} interests us in particular, as it equips ViT with the ability to use any square present in an image as a patch, removing both grid and size limitation.

\subsubsection{Active visual exploration.} 
The SLAM (simultaneous localization and mapping) challenge is often described in the context of active exploration\cite{ramakrishnan2021exploration}. 
A popular approach seen in many models\cite{ba2015multiple,elsayed2019saccader,papadopoulos2021hard,uzkent2020learning} is to feed the model a low-resolution version of the image and use a variation of a policy gradient algorithm to choose parts of an image to focus on.
Many notable works in domain of AVE are using CNN-based attention maps for glimpse selection\cite{seifi2020attend,seifi2021glimpse,seifi2019wheretolooknext,wang2020glance}. 
Simglim \cite{jha2023simglim} introduces a MAE-based model with an additional glimpse decision neural network to solve image reconstruction tasks. 
AME\cite{pardyl2023active} similarly uses MAE as a backbone, but makes decisions solely on the basis of attention maps without added loss or modules.
STAM\cite{rangrej2022consistency} uses a visual transformer and a one-step actor-critic for choosing glimpse locations in the classification task.

\section{\ours{}}
\label{sec:ours}

The key idea of Adaptive Glimpse (\ours{}) is to let the agent select both the scale and position of each successive observation (\textit{glimpse}) from a continuous action space. This way, the agent can learn to decide whether, at a given step of the exploration process, it is preferable to sample a wider view field at a lower relative resolution, zoom in on detail to capture a small high-resolution glimpse, or choose a midway solution.

\begin{figure}[t]
    \centering
    \includegraphics[scale=1,clip,trim={0 .9cm 0 .2cm}]{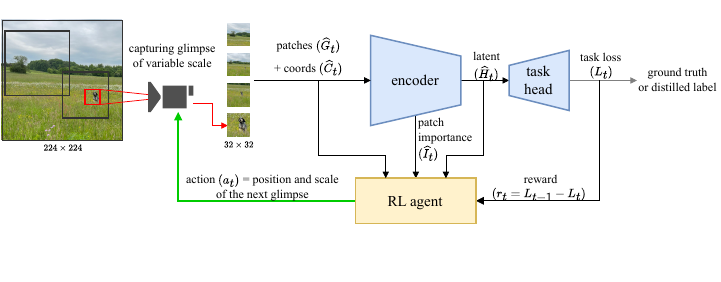}
    \vspace{-.2cm}
    \caption{\textbf{Architecture}: \ours{} consists of two parts: a vision transformer-based encoder with a task-specific head (see~\cref{sec:elasticvit}) and a Soft Actor-Critic RL agent (see~\cref{sec:rl}). At each exploration step, the RL agent selects the position and scale of the next glimpse based on the information about previous patches, their coordinates, importance, and latent representations.}
    \label{fig:arch}
\end{figure}

In this section, we start by formalizing the concept of adaptive glimpse sampling (see~\cref{sec:archoverview}), and then we discuss the main two components of \ours{} presented in~\cref{fig:arch}. In~\cref{sec:elasticvit}, we describe a vision transformer encoder with variable scale sampling and a task-specific head. In~\cref{sec:rl}, we present the reinforcement learning agent based on the Soft Actor-Critic (SAC) algorithm.

\subsection{Adaptive glimpse sampling}
\label{sec:archoverview}

\subsubsection{Glimpses.}
Let $X$ be an unobserved scene to explore. We assume $X$ to be a rectangle within the Cartesian coordinate system. A \textit{glimpse} $G$ is a square region within $X$ that can be observed by a camera, specified by the position of its top-left corner $(x, y)$ and its size $d$ (camera field of view). Furthermore, we define glimpse scale as: 
$z = (d - \dmin) / (\dmax - \dmin)$, where $\dmax$ and $\dmin$ are constants denoting the maximum and minimum field of view of an agent camera. Intuitively, a scale of $0$ corresponds to the maximum camera zoom level and a scale of $1$ to the widest view possible. Finally, let $C = (x, y, z)$ be the coordinates of glimpse $G$ and constant $\dcam \times \dcam$ denote the sampling resolution of $G$, i.e., the resolution glimpses obtained from the camera sensor.

\subsubsection{AVE process.}
Now, we can define the active visual exploration process as a sequence of glimpse selections. Let $T$ be the maximum number of exploration steps. The process starts at time $t = 0$ with an empty sequence of observations. At time $t \in \{1, ..., T\}$, the camera captures a glimpse $G_t$ at coordinates $C_t$ proposed by the model, generating a patch of resolution $\dcam \times \dcam$. We simulate the process of glimpse capturing by cropping a patch from a large image representing the environment $X$ and scaling it to $\dcam \times \dcam$. The model stores information about glimpses; therefore, at step $t$, it can access glimpses $G_1, G_2, ..., G_t$ and their corresponding coordinates. The exploration process is stopped when a set confidence level or a maximum number of glimpses is reached. As we will show in~\cref{sec:rl}, this process can be formulated as a Markov Decision Process (MDP) to leverage RL methods~\cite{sutton2018reinforcement}.

\subsection{Vision transformer with variable scale sampling}
\label{sec:elasticvit}

The architecture of our backbone network consists of two parts: an encoder based on a modified version of ViT~\cite{dosovitskiy2020vit} and a task-specific head (decoder). The goal is to perform the main task, e.g. classification or reconstruction, based on already observed glimpses while providing information for the RL agent network.

\subsubsection{Glimpse encoder.}
At each step $t$, the ViT encoder is provided with a sequence of glimpses $G_1, G_2, ..., G_{t-1}$ and their coordinates $C_1, C_2, ..., C_{t-1}$. Depending on the $\dcam$ resolution relative to the ViT native patch size $\dpatch$, each glimpse $G_i$ is divided with a standard sampling grid into a sequence of patches $G'_i = g'_{i,1}, ..., g'_{i, k}$ with coordinates $C'_i = c'_{i,1}, ..., c'_{i, k}$, where $k = (\lceil \dcam / \dpatch \rceil)^2$. The resulting sequences of patches and coordinates from all previous glimpses are concatenated into $\widehat{G}_t$ and $\widehat{C}_t$, respectively.

The standard ViT positional embeddings assume that all patches are sampled from a regular grid. As our method relaxes this constraint, we apply ElasticViT~\cite{pardyl2023beyond} positional encoding calculated according to the patch coordinates. Finally, a trainable class token is appended to the sequence, which is then passed through ViT transformer blocks. The encoder outputs a sequence of latent tokens $\widehat{H_t}$, one per patch, and the class token. Furthermore, we estimate the importance of each input patch, calculating a transformer attention rollout~\cite{abnar2020quantifying}, which results in a sequence $\widehat{I_t}$.

\subsubsection{Task-specific decoder.}
The head of the classification model is a simple linear layer taking as an input the class token, as in standard ViT. However, for dense prediction tasks (e.g., reconstruction, segmentation), a MAE-like~\cite{he2022masked} transformer decoder is used. Then, the decoder receives a sequence of all tokens from the encoder and a full grid of mask tokens as input. The mask tokens consist of a positional embedding for each position in the decoder grid and a shared learnable query embedding, indicating that the token value is to be predicted. This is in contrast to MAE, which only uses mask tokens for unknown areas of the image. However, using tokens for all positions is essential in our case because with variable scale sampling we must predict the entire image rather than solely focus on the absent portions.

The decoder consists of a series of transformer blocks, generating output tokens projected through a linear layer for reconstruction or a progressive upscale module~\cite{zheng2021rethinking} of 4 convolutional layers and interpolations for segmentation. Finally, the output mask tokens are arranged according to a grid, and the remaining ones are discarded.

\subsubsection{Training objectives.}
The entire backbone network is trained using a task-specific loss function. Optimization is performed only on the last exploration step $t=T$ after seeing all glimpses. The loss function for reconstruction is the root mean squared error (RMSE). For classification and segmentation, we use distilled soft targets computed by a teacher model and the Kullback-Leibler divergence as the loss function. The teacher model is a pre-trained ViT from~\cite{touvron2022deit} for classification and a DeepLabV3~\cite{chen2017rethinking} with a ResNet-101 backbone for segmentation. In both cases, the teacher model is provided with the entire scene $X$, as in STAM~\cite{rangrej2022consistency}.

\subsection{Soft Actor-Critic agent}
\label{sec:rl}

We consider AVE as a Markov Decision Process (MDP), where at timestep $t$, the agent observing state $s_{t}$ (information about previous glimpses) takes action $a_t$ (coordinates of the next glimpse). It leads to state $s_{t+1}$ (information about previous glimpses and the next glimpse) as well as the reward $r_{t+1}$ (scalar estimating how much the glimpse helped in refining the prediction). Presenting AVE as MDP allows us to leverage reinforcement learning algorithms.

\subsubsection{Preliminaries.}
The focal point of reinforcement learning is the policy $\pi_\theta$, which chooses the next action based on the current state, i.e., $a_t \sim \pi_\theta(s_t)$. The goal is to find the policy $\pi_\theta$ that maximizes the value function, corresponding to the expected discounted sum of rewards:
$V^{\pi_\theta}(s) := \mathbb{E}_{\pi_\theta}\left[\sum_{t=k}^{T} \gamma^t r_t \vert s_k = s\right],$
where $\gamma=0.99$ is the discount factor. The expectation is taken under the policy $\pi_\theta$, i.e., we use $\pi_\theta$ to take actions in the environment. As such, we aim to find $\theta^* = \argmax_{\theta} V^{\pi_\theta}$. Additionally, in order to evaluate actions and facilitate the learning of the policy, we define the state-action value function $Q^{\pi_\theta}(s, a) := r(s, a) + \gamma \mathbb{E}_{s'}V^{\pi_\theta}(s'),$ 
where $r(s, a)$ is the reward received in state $s$ when executing action $a$, and state $s'$ represents the next sampled state. Intuitively, it represents the expected return of our policy, that is, the value function of executing the action $a$ in the first step and choosing the subsequent actions according to the policy $\pi$. Below, we define observations, actions, rewards, the maximum-entropy objective central to the Soft Actor-Critic algorithm, and the design of our actor and critic architectures.

\begin{figure}[t]
\centering
\includegraphics[scale=0.99]{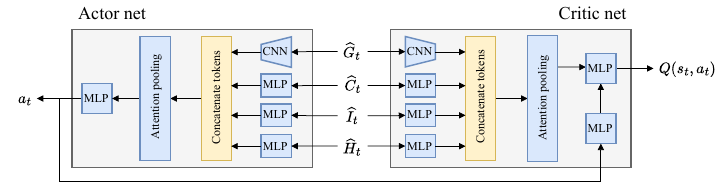}
\vspace{-.2cm}
\caption{\textbf{RL agent}: RL module of \ours{} uses two networks: the actor and the critic. The actor predicts the action $a_t$ (position and scale of the next glimpse) based on state $s_t = (\widehat{G}_t, \widehat{C}_t, \widehat{I_t}, \widehat{H_t})$. The critic estimates the $Q(s_t, a_t)$, corresponding to the expected cumulative reward for taking this action.}
\label{fig:rlagent}
\end{figure}

\subsubsection{State, action and reward.}
\label{sec:state}
To create the \textit{state} for the RL agent, we supplement the glimpses $G_t$ with additional information to make the inference easier. In particular, the state $s_{t}$ consists of sequences $(\widehat{G}_t, \widehat{C}_t, \widehat{I_t}, \widehat{H_t})$, where (as defined in~\cref{sec:archoverview} and~\ref{sec:elasticvit}):
\begin{itemize}
\item $\widehat{G}_t$ are all patches of the previously sampled glimpses, 
\item $\widehat{C}_t$ are coordinates of these patches,
\item $\widehat{I_t}$ are their importance (one importance value per patch),
\item $\widehat{H_t}$ are latent representations of these patches (one token per patch).
\end{itemize}
Notice that the starting state $s_0$ is an empty sequence corresponding to the fact that we do not know anything about the environment.

\ours{} proposes continuous-valued \textit{actions} that describe the arbitrary position and scale of an image. Therefore, the action is a tuple $(x, y, z) \in [0, 1]^3$, where $x, y$ represents the normalized coordinates of the top-left glimpse corner and $z$ is its scale.

Finally, we define the \textit{reward} as the difference between the loss in the successive timesteps, i.e. $r_t = L_{t-1} - L_t$.

\subsubsection{Soft Actor-Critic.}
RL objective can be optimized with various approaches~\cite{sutton2018reinforcement}. However, since exploration is crucial to solving AVE, we decided to use Soft Actor-Critic~\cite{haarnoja2018soft} (SAC) reinforcement learning algorithm. SAC operates in the maximum entropy framework, meaning that besides maximizing the expected sum of rewards, it also takes into account the entropy of the action distribution. That is, the goal is to optimize 
$V^{\pi_\theta}(s) := \mathbb{E}_{\pi}\left[\sum_{t=k}^{T} \gamma^t r_t + \alpha\ent(\pi(s_t)) |s_k = s\right],$
where $\ent$ is the entropy, and $\alpha$ is used to weigh its importance. Higher $\alpha$ values encourage the RL algorithm to find more exploratory policies, resulting in more diverse actions.

\subsubsection{Actor and Critic architectures.}
\ours{} requires two networks: the actor that encodes the policy $\pi$ and the critic that encodes the state-action value function $Q$. For this purpose, we build a custom-crafted architecture, see~\cref{fig:rlagent}. In particular, we create separate token encoders for each part of the input $s_t$: a small convolutional network that processes patches in $\widehat{G}_t$, and a small MLP for each of the other inputs $\widehat{C}_t, \widehat{I_t}, \widehat{H_t}$. As a result, we obtain four embedding vectors for each patch. We concatenate and process them using an attention pooling~\cite{ilse2018attention} layer to combine information across patches. Finally, we use another MLP to obtain the action $a_{t}$ for the actor and the value prediction $Q(s_t, a_t)$ for the critic. Despite the similarity in the actor and critic architectures, we do not share any parameters between them as it destabilizes the training process.

\section{Experimental Setup}
\label{sec:setup}

\subsubsection{Architecture.}
In all our experiments we use an encoder of the same size as standard ViT-B~\cite{dosovitskiy2020vit}, i.e., 12 transformer blocks and embedding size of 768. The decoder consists of 8 blocks with the embedding size of 512. For the RL networks the hidden dimension size is 256, the number of attention heads is 8 and MLPs have 3 layers. All networks use GELU activation functions~\cite{hendrycks2016gaussian}.

\subsubsection{Training.}
We adopt the AdamW optimization algorithm~\cite{loshchilov2018decoupled}, setting the weight decay value to $10^{-4}$ and the initial learning rate to $10^{-5}$ for classification and $10^{-4}$ for other tasks. The learning rate is then decayed using a half-cycle cosine rate to $10^{-8}$ for the remainder of the training. The model is trained for 100 epochs with early stopping. During training, we alternate between optimizing the backbone and the RL agent each epoch, except for the first 30 epochs when we train the RL agent only. We augment the training data using the 3-Augment regime proposed in~\cite{touvron2022deit} extended with a random affine transform for the segmentation target. Additionally, we pre-train the model for 600 epochs with 196 random glimpses per image with sizes and positions sampled from a uniform distribution. In segmentation experiments we fine-tune a model trained for reconstruction to accommodate for the relatively small size of the dataset.

\begin{figure}[t]
\begin{center}
\begin{small}

\begin{tabular}{r@{\hskip2pt}c@{\hskip2pt}c@{\hskip2pt}c@{\hskip2pt}c@{\hskip2pt}c@{\hskip2pt}|@{\hskip2pt}c@{\hskip2pt}c}
A) & \tabfigure{width=\smallerimg}{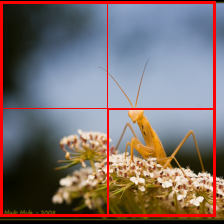} & \tabfigure{width=\smallerimg}{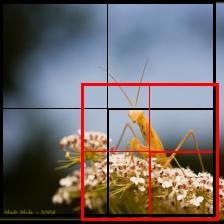} & \tabfigure{width=\smallerimg}{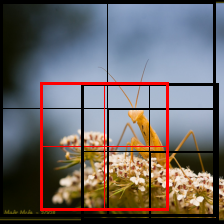} & \tabfigure{width=\smallerimg}{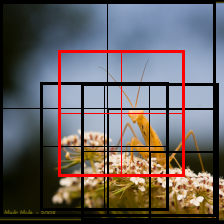} & \tabfigure{width=\smallerimg}{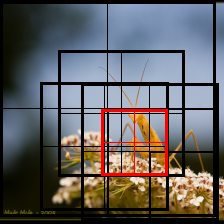} & \tabfigure{width=\smallerimg}{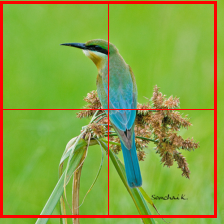} & \tabfigure{width=\smallerimg}{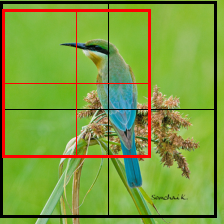} \\[19pt]
B) & \tabfigure{width=\smallerimg}{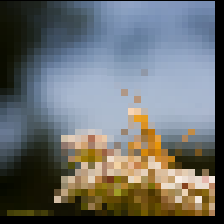} & \tabfigure{width=\smallerimg}{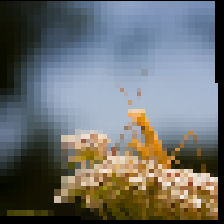} & \tabfigure{width=\smallerimg}{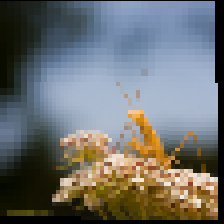} & \tabfigure{width=\smallerimg}{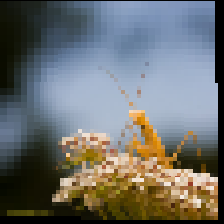} & \tabfigure{width=\smallerimg}{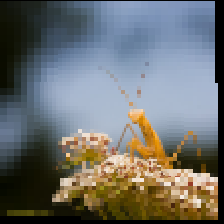} & \tabfigure{width=\smallerimg}{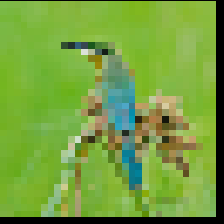} & \tabfigure{width=\smallerimg}{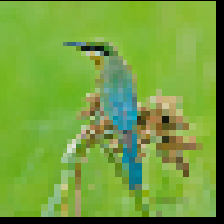} \\[19pt]
C) & snail & hopper & agama & hopper & {\color{forestgreen4416044}mantis} & weevil & {\color{forestgreen4416044}bee eater}  \\[2pt]
D) & 22\% & 33\% & 49\% & 36\% & {\color{forestgreen4416044}76\%} & 26\% & {\color{forestgreen4416044}88\%}
\end{tabular}
\end{small}
\end{center}
\vspace{-.2cm}
\caption{\textbf{Glimpse selection step-by-step:} \ours{} explores $224 \times 224$ images from ImageNet with $32 \times 32$ glimpses of variable scale, zooming in on objects of interest and stopping the process after reaching $75\%$ predicted probability. The rows correspond to: A) glimpse locations, B) pixels visible to the model (interpolated from glimpses for preview), C) predicted label, D) prediction probability.}
\label{fig:qualiative-cls}
\end{figure}

\subsubsection{Datasets and metrics.}
We assess our method on several publicly available vision datasets. ImageNet-1k~\cite{deng2009imagenet} is used for classification and reconstruction tasks. The performance of zero-shot reconstruction is evaluated on MS COCO 2014~\cite{lin2014microsoft}, ADE20K~\cite{zhou2017scene} and SUN360~\cite{song2015sun}. Semantic segmentation is evaluated on the ADE20K dataset (the MIT scene parsing benchmark subset)~\cite{zhou2017scene}. Since the SUN360 dataset does not have a predetermined train-test split, we use a 9:1 train-test split according to an index provided by the authors of \cite{seifi2021glimpse}. We report accuracy for classification tasks, root mean squared error (RMSE) for reconstruction, and pixel average precision (AP), class-mean average precision (mAP) and class-mean intersection over union score (mIoU) for segmentation.

\subsubsection{Glimpse regimes} \label{sec:glimpseregimes} We compare our model to baselines with different glimpse regimes, that can be categorized as follows: \begin{enumerate*}[label=(\alph*)]
    \item simple - square glimpses with a fixed and constant resolution~\cite{rangrej2022consistency, pardyl2023active, jha2023simglim},
    \item retinal - retina-like glimpses\cite{sandini2003retina} with more pixels in the center than on the edges~\cite{seifi2020attend, seifi2021glimpse, pardyl2023active},
    \item full+simple - one low-resolution glimpse of the entire scene followed by simple glimpses~\cite{ba2015multiple,wang2020glance,elsayed2019saccader,uzkent2020learning,recasens2018learning,papadopoulos2021hard}, and
    \item adaptive - our variable scale glimpse regime.
\end{enumerate*}

For easy comparison of different glimpse regimes, we provide a \textit{pixel percentage} metric representing the percentage of image pixels known to the model, as defined in~\cite{pardyl2023active}, which is calculated as the number of pixels captured in all glimpses divided by the number of pixels in the full scene image.

\section{Results}

\begin{table}[t]
\caption{\textbf{Reconstruction results:} RMSE (lower is better) obtained by our model for reconstruction task against AttSeg~\protect\cite{seifi2020attend}, GlAtEx~\protect\cite{seifi2021glimpse}, SimGlim~\protect\cite{jha2023simglim}, and AME~\cite{pardyl2023active} on ImageNet-1k, SUN360, ADE20K and MS COCO datasets. Regardless of the number of glimpses, as well as their resolution and regime (see~\cref{sec:glimpseregimes}), our method outperforms competitive solutions. Note that Pixel \% denotes the percentage of image pixels known to the model, $\dag$ a reproduced result not published in the relevant paper, and * zero-shot performance.}
\label{tab:reconstruction}
\begin{center}
\begin{small}
\vspace{-.2cm}
\begin{tabular}{l@{\hskip6pt}lllllr@{\hskip8pt}cr}
\toprule
Method & IMNET & SUN360 & ADE20k & COCO & Image res. & Glimpses & Regime & Pixel \% \\
\midrule
AME & 30.3$^\dag$ & 29.8 & 30.8 & 32.5 & $128 \times 256$ & $8 \times 32^2$ &simple & 25.00 \\
\textbf{Ours} & \textbf{14.5} & \textbf{11.1}* & \textbf{14.0}* & \textbf{14.5}* & $224 \times 224$ & $12 \times 32^2$ & adaptive & 24.49 \\
\midrule
AttSeg &-- & 37.6 & 36.6 & 41.8 & $128 \times 256$ & $8 \times 48^2$ &retinal & 18.75 \\
GlAtEx & --& 33.8 & 41.9 & 40.3 & $128 \times 256$ & $8 \times 48^2$ &retinal & 18.75 \\
AME & --& 23.6 & 23.8 & 25.2 & $128 \times 256$ & $8 \times 48^2$ & retinal & 18.75 \\
SimGlim &-- & 26.2 & 27.2 & 29.8 & $224 \times 224$ & $37 \times 16^2$ &simple & 18.75 \\
AME & --& 23.4 & 26.2 & 28.6 & $224 \times 224$ & $37 \times 16^2$ & simple & 18.75 \\
\textbf{Ours} & \textbf{14.7} & \textbf{11.1}* & \textbf{14.2}* & \textbf{14.7}* & $224 \times 224$ & $9 \times 32^2$ & adaptive & 18.36 \\
\midrule
AME & --& 37.9 & 40.7 & 43.2 & $128 \times 256$ & $8 \times 16^2$ & simple & 6.25 \\
\textbf{Ours} & \textbf{20.9} & \textbf{17.6}* & \textbf{20.5}* & \textbf{21.5}* & $224 \times 224$ & $12 \times 16^2$ & adaptive & 6.12 \\
\textbf{Ours} & \textbf{20.7} & \textbf{17.2}* & \textbf{20.7}* & \textbf{21.4}* & $224 \times 224$ & $3 \times 32^2$ & adaptive & 6.12 \\
\bottomrule
\end{tabular}
\end{small}
\end{center}
\end{table}
\begin{figure}[th!]
\begin{center}
\begin{small}
\begin{sc}
\begin{tabular}{c@{\hskip2pt}c@{\hskip2pt}c@{\hskip2pt}c@{\hskip2pt}c}
Input&Ours&AME&GlAtEx&AttSeg\\
\includegraphics[width=\smallimg]{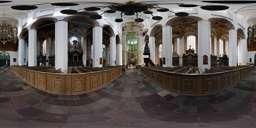}&\includegraphics[width=\smallimg]
{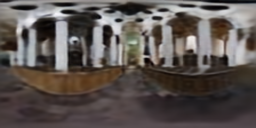}&\includegraphics[width=\smallimg]
{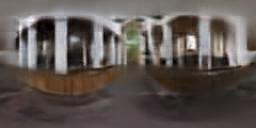}&\includegraphics[width=\smallimg]{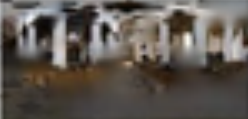}&\includegraphics[width=\smallimg]{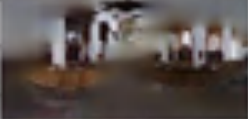}\\
\includegraphics[width=\smallimg]{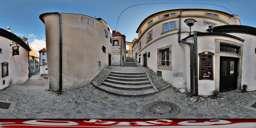}&\includegraphics[width=\smallimg]{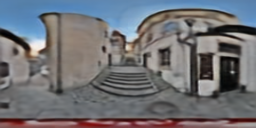}&\includegraphics[width=\smallimg]
{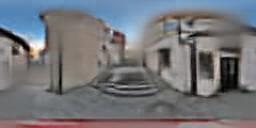}&\includegraphics[width=\smallimg]{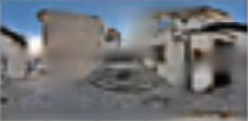}&\includegraphics[width=\smallimg]{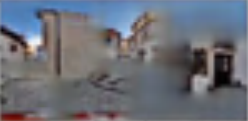}\\
\end{tabular}
\begin{tabular}{c@{\hskip2pt}c@{\hskip2pt}c@{\hskip2pt}c|c@{\hskip2pt}c@{\hskip2pt}c@{\hskip2pt}c}
Input&Ours&AME&SimGlim&Input&Ours&AME&SimGlim\\
\includegraphics[width=\smallerimg]{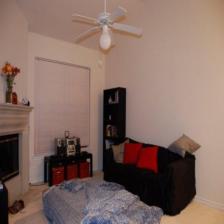}&\includegraphics[width=\smallerimg]{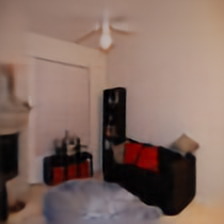}&\includegraphics[width=\smallerimg]{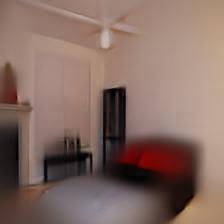}&\includegraphics[width=\smallerimg]{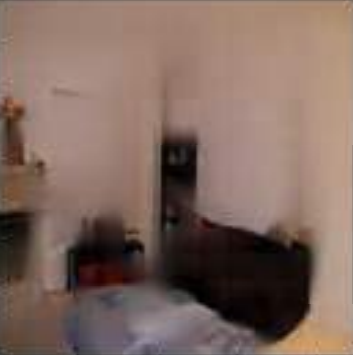}&
\includegraphics[width=\smallerimg]{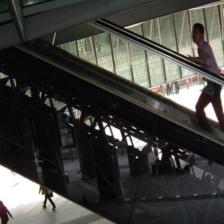}&\includegraphics[width=\smallerimg]{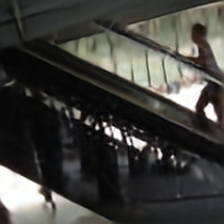}&\includegraphics[width=\smallerimg]{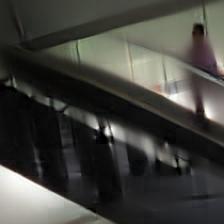}&\includegraphics[width=\smallerimg]{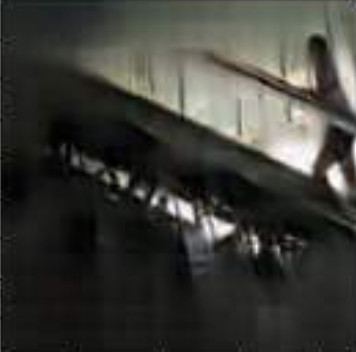}\\
\includegraphics[width=\smallerimg]{}&\includegraphics[width=\smallerimg]{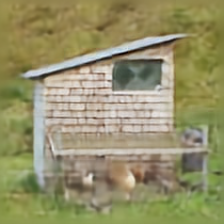}&\includegraphics[width=\smallerimg]{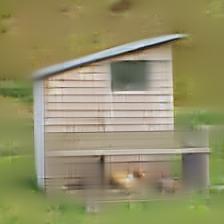}&\includegraphics[width=\smallerimg]{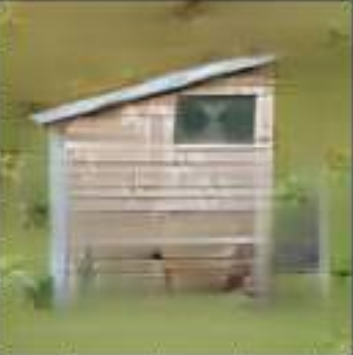}&
\includegraphics[width=\smallerimg]{}&\includegraphics[width=\smallerimg]{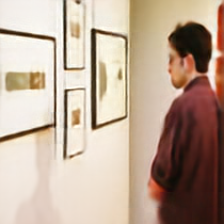}&\includegraphics[width=\smallerimg]{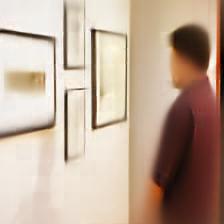}&\includegraphics[width=\smallerimg]{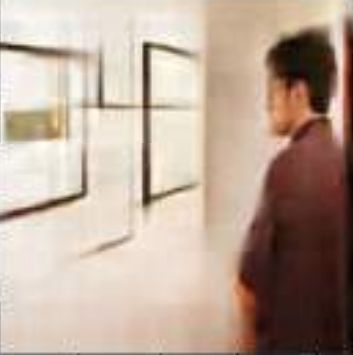}\\
\end{tabular}
\end{sc}
\end{small}
\end{center}
\vspace{-.2cm}
\caption{\textbf{Reconstruction quality for SUN360 (top) and ADE20K (bottom):} Sample reconstructions of our method compared with AME \cite{pardyl2023active}, AttSeg~\protect\cite{seifi2020attend}, GlAtEx~\protect\cite{seifi2021glimpse} and SimGlim~\protect\cite{jha2023simglim} on the SUN360 and ADE20K datasets. Reconstructions done with our method are visibly more detailed and less blurry than those obtained by baseline methods. Notice that images for comparison were taken from the baseline publications (we did not select them).}
\label{fig:qualiative}
\end{figure}
\begin{table}[t]
  \caption{\textbf{Classification results}: Accuracy obtained by our model for classification task against DRAM~\cite{ba2015multiple}, GFNet~\cite{wang2020glance}, Saccader~\cite{elsayed2019saccader}, STN~\cite{recasens2018learning}, TNet~\cite{papadopoulos2021hard}, PatchDrop~\cite{uzkent2020learning} and STAM~\cite{rangrej2022consistency} on ImageNet-1k dataset. Our \ours{} needs 40\% less pixels to match the performance of the best baseline method. Note that Pixel \% denotes the percentage of image pixels known to the model, while regimes are described in~\cref{sec:glimpseregimes}.}
  \label{tab:classification}
  \vspace{-.2cm}
\begin{center}
\begin{small}
  \begin{tabular}{l@{\hskip12pt}l@{\hskip12pt}r@{\hskip12pt}c@{\hskip12pt}r} 
    \toprule
    Method & Accuracy \% & Glimpses & Regime & Pixel \%  \\ 
    \midrule
    DRAM & 67.50 & $8 \times 77^2$ & full+simple & 94.53 \\
    GFNet & 75.93 & $5 \times 96^2$ & full+simple & 91.84 \\
    Saccader & 70.31 & $6 \times 77^2$ & full+simple & 70.90 \\
    TNet & 74.62 & $6 \times 77^2$ & full+simple & 70.90 \\
    STN & 71.40 &  $9 \times 56^2$& full+simple & 56.25 \\
    PatchDrop & 76.00 &  $\sim8.9 \times 56^2$ & full+simple+stopping & $\sim$55.63 \\
    STAM & 76.13 & $14 \times 32^2$ & simple & 28.57\\
    \textbf{Ours} & \textbf{77.54} & $14 \times 32^2$ & adaptive & 28.57\\
    \textbf{Ours} & \textbf{76.30} &  $\sim8.3 \times 32^2$ & adaptive+stopping & $\sim$\textbf{16.94}\\
    \bottomrule
  \end{tabular}
\end{small}
\end{center}
\end{table}

\begin{table}[th]
\caption{\textbf{Segmentation results:} Comparison of our model against AttSeg~\protect\cite{seifi2020attend}, GlAtEx~\protect\cite{seifi2021glimpse}, and AME~\cite{pardyl2023active} on the ADE20K dataset. Our method performs on pair with AME, but requires $35\%$ less pixels, while outperforming other competitive methods on all considered metrics: Pixel-wise Accuracy (mPA, higher is better), Pixel-Accuracy (PA, higher is better), and Intersection over Union (IoU, higher is better).}
\vspace{-.2cm}
\label{tab:segmentation}
\begin{center}
\begin{small}
\begin{tabular}{l@{\hskip12pt}l@{\hskip12pt}l@{\hskip12pt}l@{\hskip12pt}l@{\hskip12pt}r@{\hskip12pt}c@{\hskip12pt}r}
\toprule
Method  & PA \% & mPA \% & IoU \%  & Image res. & Glimpses & Regime & Pixel \% \\
\midrule
AttSeg & 47.9 & -- & -- & $128 \times 256$ & $8 \times 48^2$ &retinal & 18.75  \\
GlAtEx & 52.4 & -- & -- & $128 \times 256$ & $8 \times 48^2$ & retinal & 18.75  \\
Ours & \textbf{67.4} & \textbf{29.4} & \textbf{22.7} & $224 \times 224$ & $4 \times 48^2$ & adaptive & 18.36 \\
\midrule
AME & 70.3 & 32.2 & 24.4 & $128 \times 256$ & $8 \times 48^2$ &simple & 56.25 \\
Ours & 70.0 & 32.8 & 25.7 & $224 \times 224$ & $8 \times 48^2$ & adaptive & \textbf{36.73}  \\
\bottomrule
\end{tabular}
\end{small}
\end{center}
\end{table}

\begin{figure}[t]
\begin{center}
\begin{small}
\begin{tabular}{c@{\hskip2pt}c@{\hskip2pt}c@{\hskip2pt}c@{\hskip2pt}c}
Image&Target&Ours&AME&GlAtEx\\
\includegraphics[width=\smallestimg,height=\smallestimg]{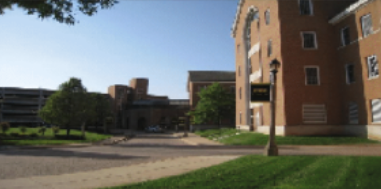}&\includegraphics[width=\smallestimg,height=\smallestimg]{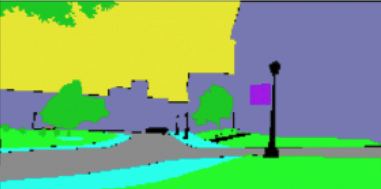}&\includegraphics[width=\smallestimg,height=\smallestimg]{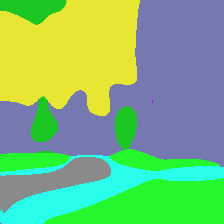}&\includegraphics[width=\smallestimg,height=\smallestimg]{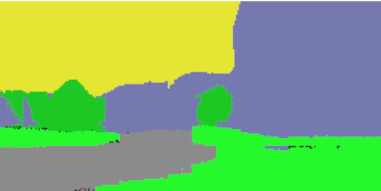}&\includegraphics[width=\smallestimg,height=\smallestimg]{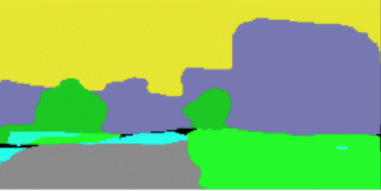}\\
\includegraphics[width=\smallestimg,height=\smallestimg]{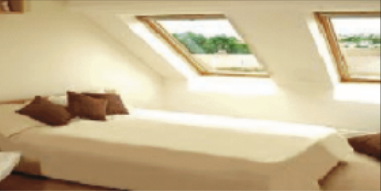}&\includegraphics[width=\smallestimg,height=\smallestimg]{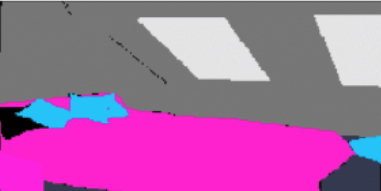}&\includegraphics[width=\smallestimg,height=\smallestimg]{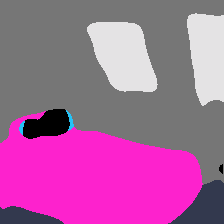}&\includegraphics[width=\smallestimg,height=\smallestimg]{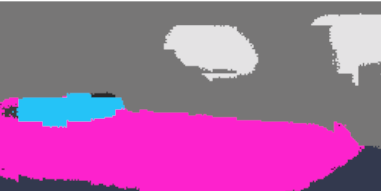}&\includegraphics[width=\smallestimg,height=\smallestimg]{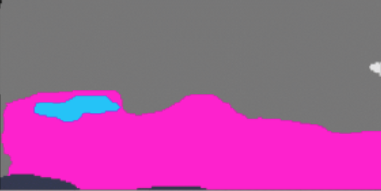}\\
\includegraphics[width=\smallestimg,height=\smallestimg]{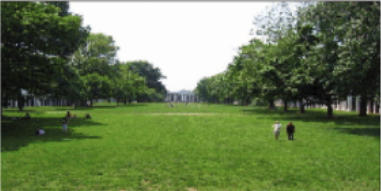}&\includegraphics[width=\smallestimg,height=\smallestimg]{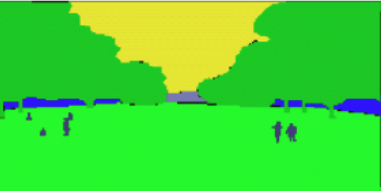}&\includegraphics[width=\smallestimg,height=\smallestimg]{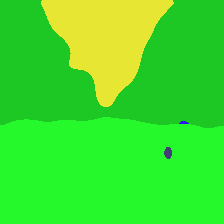}&\includegraphics[width=\smallestimg,height=\smallestimg]{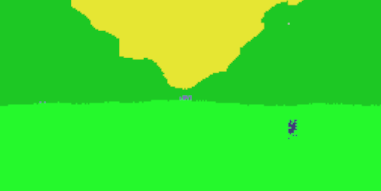}&\includegraphics[width=\smallestimg,height=\smallestimg]{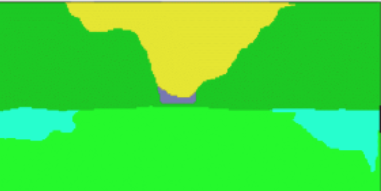}\\
\includegraphics[width=\smallestimg,height=\smallestimg]{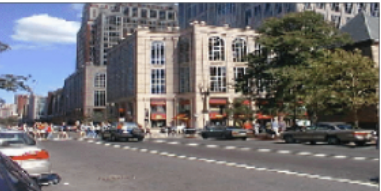}&\includegraphics[width=\smallestimg,height=\smallestimg]{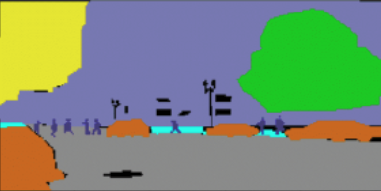}&\includegraphics[width=\smallestimg,height=\smallestimg]{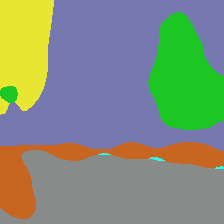}&\includegraphics[width=\smallestimg,height=\smallestimg]{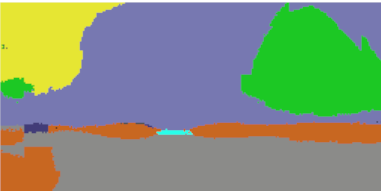}&\includegraphics[width=\smallestimg,height=\smallestimg]{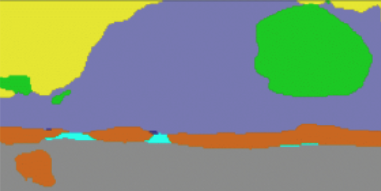}\\
\end{tabular}
\end{small}
\caption{\textbf{Segmentation qualitative results.} Sample semantic segmentation of our method compared with AME~\cite{pardyl2023active} and GlAtEx~\protect\cite{seifi2021glimpse} on the ADE20k dataset. In terms of quality, the segmentation maps generated by our approach are at least comparable to those produced by competing methods.}
\end{center}
\label{fig:qualiative-seg}
\end{figure}

In this section, we present an evaluation of \ours{} compared to competitive methods, followed by an analysis of our approach. Both quantitative and qualitative results for all baseline methods were taken from~\cite{pardyl2023active} for reconstruction and segmentation, and~\cite{rangrej2022consistency} for classification. Further results are provided in the supplementary materials.

An overview of glimpse selection performed by our model is portrayed in \cref{fig:qualiative-cls}. The visualization demonstrates the arbitrary glimpse position and scale capabilities of our method. \ours{} detects objects of interest and zooms in on them to extract fine details.

\subsection{Tasks}

\subsubsection{Reconstruction.}

Reconstructing the entire scene from observed glimpses verifies comprehensive scene understanding. In~\cref{tab:reconstruction} we group the results by pixel percentage (fraction of pixels of the original input image known to the model). Our approach outperforms existing methods by a large margin. In particular, with only 6\% of the pixels seen, \ours{} performs better than other methods that had over 18\% of the image visible to them. Note that unlike baseline methods, we train our model only on ImageNet-1k without fine-tuning for each evaluated dataset. Qualitative outcomes in~\cref{fig:qualiative} showcase \ours{} producing reconstructions with a higher level of detail, reproducing more objects compared to baseline methods. All models except AttSeg have backbone networks pre-trained on ImageNet-1k; AttSeg's pre-training details were not disclosed.

\subsubsection{Classification.}

Results for multi-class classification on the ImageNet-1k dataset can be found in~\cref{tab:classification}. \ours{} outperforms all prior methods, achieving $77.54\%$ $(\pm 0.18)$ accuracy compared to $76.13\%$ of the best baseline method STAM~\cite{rangrej2022consistency}. With early exploration termination after reaching $85\%$ probability of predicted class, it requires over $40\%$ less pixels to match STAM accuracy.
Visualizations of glimpse selection for classification are presented in~\cref{fig:qualiative-cls}. With an early stopping probability threshold of $75\%$, it is able to classify $224\times224$ images using only a few $32\times32$ glimpses of $4$ patches each.

\subsubsection{Segmentation.}

The goal of the semantic segmentation task is to classify each pixel of the full scene based on the captured glimpses. The numerical results are presented in~\cref{tab:segmentation}. \ours{} outperforms both AttSeg~\cite{seifi2020attend} and GlAtEx~\cite{seifi2021glimpse} by a large margin. It performs on par with AME~\cite{pardyl2023active} in terms of accuracy; however, it requires $35\%$ less information to achieve this result. Consistently, visualizations in~\cref{fig:qualiative-seg} confirm the quality of the produced segmentation maps.

\subsection{Analysis}

\begin{figure}[th!]
\begin{center}
\begin{small}
\begin{subfigure}[b]{.45\textwidth}
\begin{tikzpicture}
\tikzstyle{every node}=[font=\scriptsize]
\begin{axis}[
    title={SUN360 reconstruction},
    xlabel={Pixel \%},
    ylabel={Root-mean-square error},
    xtick={0,5,10,15,20,25},
    xmin=0,
    xmax=25,
    ymajorgrids=true,
    grid style=dashed,
    scale only axis,
    height=3cm,
    ymax=85,
    width=.7\linewidth,
    legend cell align={left}
]
\addplot[color=crimson2143940]
    coordinates {
    (2.34,68.99)(4.68,62.32)(7.03,56.24)(9.375,49.57)(11.71,46.09)(14.06,41.49)(16.40,37.49)(18.75,35.71)(21.09,34.66)(23.43,31.58)
    };
\addlegendentry{AttSeg}
\addplot[color=forestgreen4416044]
    coordinates {
    (2.34,70.50)(4.68,59.81)(7.03,51.77)(9.375,46.25)(11.71,42.45)(14.06,40.31)(16.40,37.16)(18.75,34.72)(21.09,32.00)(23.43,31.58)
    };
\addlegendentry{GlAtEx}
\addplot[color=darkorange25512714]
    coordinates {
    (2.34,46.89)(4.68,40.11)(7.03,35.40)(9.375,31.99)(11.71,28.99)(14.06,26.93)(16.40,25.23)(18.75,23.69)(21.09,22.49)(23.43,21.56)
    };
\addlegendentry{AME}
\addplot[color=steelblue31119180]
    coordinates {
    (2.04,25.17)(4.08,19.73)(6.12,17.21)(8.16,14.45)(10.20,12.58)(12.24,11.92)(14.28,11.54)(16.32,11.34)(18.36,11.22)(20.40,11.16)(22.44,11.11)(24.48,11.12)
    };
\addlegendentry{Ours}
\end{axis}
\end{tikzpicture}
\end{subfigure}
\begin{subfigure}[b]{.45\textwidth}
\begin{tikzpicture}
\tikzstyle{every node}=[font=\scriptsize]
\begin{axis}[
    title={ImageNet-1k classification},
    xlabel={Pixel \%},
    ylabel={Accuracy \%},
    xtick={0,5,10,15,20,25,30},
    ymajorgrids=true,
    grid style=dashed,
    scale only axis,
    height=3cm,
    legend pos=south east,
    width=.7\textwidth,
    legend cell align={left}
]
\addplot[color=red]
    coordinates {
(2.04,14.52)
(4.08,35.64)
(6.12,47.64)
(8.16,55.56)
(10.20,60.91)
(12.24,64.68)
(14.28,67.47)
(16.32,69.66)
(18.36,71.25)
(20.40,72.60)
(22.44,73.73)
(24.48,74.64)
(26.53,75.47)
(28.57,76.13)
    };
\addlegendentry{STAM}
\addplot[color=steelblue31119180]
coordinates {
(2.04,29.57)
(4.08,49.69)
(6.12,58.64)
(8.16,64.04)
(10.20,67.89)
(12.24,70.76)
(14.28,72.61)
(16.32,74.12)
(18.36,75.27)
(20.40,75.94)
(22.44,76.54)
(24.48,76.96)
(26.53,77.44)
(28.57,77.54)
};
\addlegendentry{Ours}
\addplot[color=mediumpurple148103189]
coordinates {
(9.87,67.21) 
(10.89,69.66) 
(12.06,71.62) 
(13.26,73.64) 
(14.87,74.94) 
(16.93,76.30) 
(21.18,77.15) 
(28.14,77.45) 
};
\addlegendentry{Ours (stopping)}
\end{axis}
\end{tikzpicture}
\end{subfigure}
\end{small}
\end{center}
\vspace{-.2cm}
\caption{\textbf{Percentage of image pixels observed}: Figures present the relationship between the amount of pixels observed by the model relative to the full scene resolution (pixel \%), and its performance. \ours{} outperforms competitive solutions, requiring significantly less information to achieve the same performance level.}
\label{fig:nglimpses}
\end{figure}
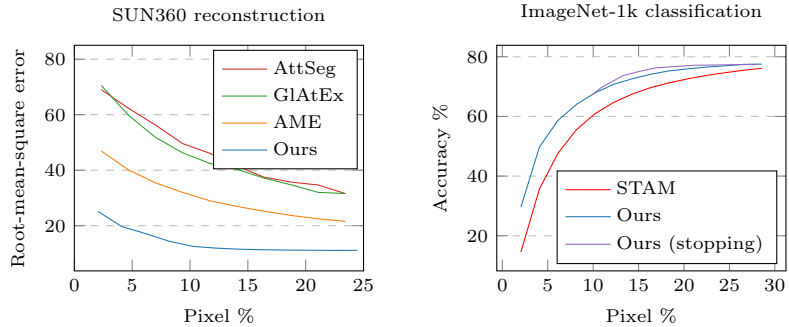


\subsubsection{Percentage of image pixels.} The relationship between the percentage of the full image pixels known to the model (pixel \%, see~\cref{sec:glimpseregimes}) and performance is plotted in~\cref{fig:nglimpses}. \ours{} requires fewer pixels to perform better than the baseline methods. In particular for reconstruction, with only 5\% of pixels it produces superior results to those achieved by competitive approaches when provided with 25\% of scene pixels.

\begin{table}[t]
\caption{\textbf{Importance of state components}: The RL state consists of the sequence $(\widehat{G}_t, \widehat{C}_t, \widehat{I_t}, \widehat{H_t})$ as described in~\cref{sec:state}. For this study, we replaced each element with its mean value (averaged over the entire dataset) to see how important it is for the model. As a result, we observe that the transformer latent is the most informative part of the state, followed by glimpse coordinates.}
\vspace{-.2cm}
\label{tab:state_ablation}
\begin{center}
\begin{small}
\begin{tabular}{c@{\hskip12pt}c@{\hskip12pt}c@{\hskip12pt}c@{\hskip12pt}r}
\toprule
patches $\widehat{G}_t$ & coordinates $\widehat{C}_t$ & importance $\widehat{I}_t$ & latent $\widehat{H}_t$ & Accuracy \% \\
\midrule
\cmarkg & \cmarkg & \cmarkg & \cmarkg & 77.54 \\
\xmark & \cmarkg & \cmarkg & \cmarkg & 76.99 \\
\cmarkg & \xmark & \cmarkg & \cmarkg & 68.25 \\
\cmarkg & \cmarkg & \xmark & \cmarkg & 77.36 \\
\cmarkg & \cmarkg & \cmarkg & \xmark & 61.82 \\
\bottomrule
\end{tabular}
\end{small}
\end{center}
\end{table}

\subsubsection{Importance of RL state elements.}
\label{sec:ablation}
The key component of \ours{} reinforcement learning algorithm is the state, which consists of the sequence $(\widehat{G}_t, \widehat{C}_t, \widehat{I_t}, \widehat{H_t})$ as described in~\cref{sec:state}. In~\cref{tab:state_ablation}, we present the ImageNet-1k classification performance when omitting and replacing each component with its mean value during inference. The resulting lower accuracy proves the significance of each component. The transformer latent $\widehat{H_t}$ and glimpse positions $\widehat{C}_t$ are especially crucial for this task, highlighting both the importance of the glimpse location within the original scene and the benefit of using the processed input over the original image.

\begin{figure}[th!]
\begin{center}
\begin{small}
\begin{tabular}{c@{\hskip10pt}c@{\hskip2pt}c@{\hskip2pt}c@{\hskip2pt}c@{\hskip2pt}c@{\hskip2pt}c@{\hskip2pt}c@{\hskip2pt}c}
\multicolumn{9}{l}{ImageNet-1k reconstruction, 16x16 glimpses}\\
 \tabfigure{width=\smallestimg}{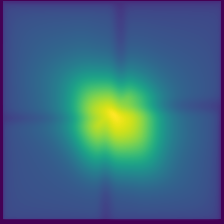} & \tabfigure{width=\smallestimg}{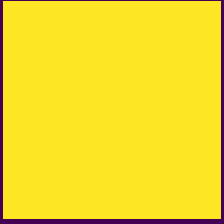} & \tabfigure{width=\smallestimg}{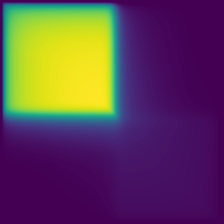} & \tabfigure{width=\smallestimg}{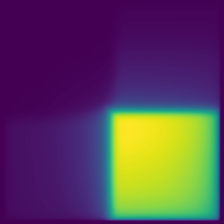} & \tabfigure{width=\smallestimg}{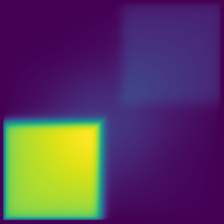} & \tabfigure{width=\smallestimg}{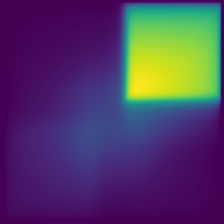} & \tabfigure{width=\smallestimg}{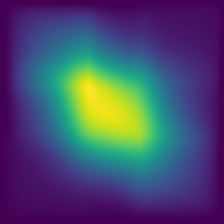} & \tabfigure{width=\smallestimg}{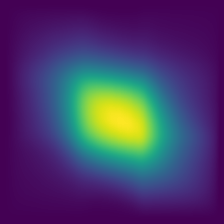} & \tabfigure{width=\smallestimg}{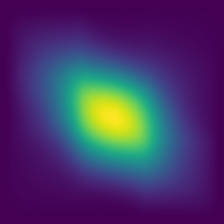} \\[18pt]
 \multicolumn{9}{l}{ImageNet-1k classification, 32x32 glimpses}\\
 \tabfigure{width=\smallestimg}{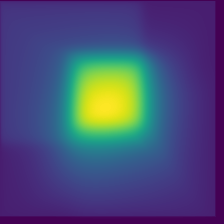} & \tabfigure{width=\smallestimg}{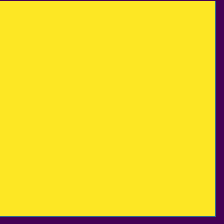} & \tabfigure{width=\smallestimg}{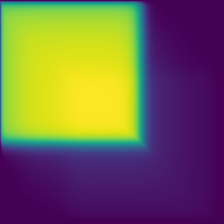} & \tabfigure{width=\smallestimg}{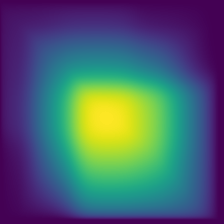} & \tabfigure{width=\smallestimg}{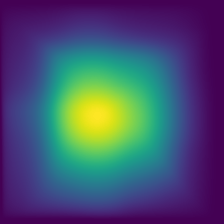} & \tabfigure{width=\smallestimg}{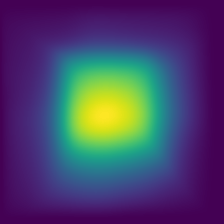} & \tabfigure{width=\smallestimg}{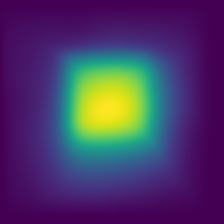} & \tabfigure{width=\smallestimg}{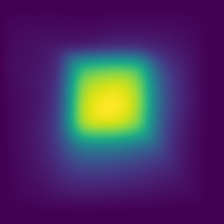} & \tabfigure{width=\smallestimg}{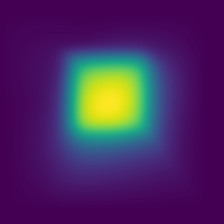} \\[16pt]
 avg. & 1 & 2 & 3 & 4 & 5 & 6 & 7 & 8 \\
\end{tabular}
\end{small}
\end{center}
\vspace{-.2cm}
\caption{\textbf{Average glimpse image:} Mean glimpse maps for models trained for reconstruction (top) and classification (bottom) averaged over all test images. On the left, an average map for all glimpses is presented, followed by maps for successive glimpses $t={1, ..., 8}$. One can observe that \ours{} learns to select the entire image as the first glimpse for both tasks, but subsequent glimpse maps differ. Four successive glimpses in reconstruction concentrate on four parts of the image, while for classification, they mostly explore the center.}
\label{fig:avgglimpse}
\end{figure}

\subsubsection{Glimpse location.}
In~\cref{fig:avgglimpse}, we illustrate the average location of each subsequent glimpse, revealing a notable distinction between reconstruction and classification tasks explored by \ours{}. In the reconstruction task, attention spans all image regions initially and later focuses on key elements. In contrast, in the classification task our model swiftly identifies crucial class-specific elements in the image center, directing attention those regions early on. Notably, both tasks uncover the well-known fact about ImageNet-1k, where key objects are concentrated around the image center~\cite{DBLP:journals/corr/KummererTB14}.

\section{Discussion and Conclusions}

This paper presents \ours{}, a novel approach to Active Visual Exploration, which enables the selection and processing of glimpses at arbitrary positions and scales. We formulate the glimpse selection problem as a Markov Decision Process with continuous action space and leverage the Soft Actor-Critic reinforcement learning algorithm, which specializes in exploration problems. Our task-agnostic architecture allows for a more efficient exploration and understanding of environments, significantly reducing the number of observations needed. \ours{} can quickly analyze the scene with large low-resolution glimpses before zooming in on details for a closer inspection. Its success across multiple benchmarks suggests a broad applicability and potential for further development in embodied AI and robotics.

While excelling in exploration, \ours{} is limited in performance by the underlying transformer architecture, which incurs quadratic computational cost relative to the number of sampled patches. A possible way to overcome this limitation is to replace it with a selective structured state-space model~\cite{gu2021efficiently,gu2023mamba}. Finally, although \ours{} perform well on current benchmarks, they do not fully reflect the complexity of Active Visual Exploration, as they do not incorporate dynamic scenes that change over time. As such, further evaluations are required before real-life deployment.

\section*{Acknowledgments}

This paper has been supported by the Horizon Europe Programme (HORIZON-CL4-2022-HUMAN-02) under the project "ELIAS: European Lighthouse of AI for Sustainability", GA no. 101120237.
This research was funded by National Science Centre, Poland (grant no. 2023/49/N/ST6/02465, 2022/47/B/ST6/03397, 2022/45/B/ST6/02817, and 2023/50/E/ST6/00469).
The research was supported by a grant from the Faculty of Mathematics and Computer Science under the Strategic Programme Excellence Initiative at Jagiellonian University.
We gratefully acknowledge Polish high-performance computing infrastructure PLGrid (HPC Center: ACK Cyfronet AGH) for providing computer facilities and support within computational grant no. PLG/2023/016558.


%
%
\bibliographystyle{splncs04}
\bibliography{main}
\end{document}


\title{\ours{}: Active Visual Exploration with Arbitrary Glimpse Position and Scale -- Supplementary Materials} 

\titlerunning{\ours{}}

\author{Adam Pardyl\inst{1,2,3}\orcidlink{0000-0002-3406-6732} \and %
Michał Wronka\inst{2}\orcidlink{0009-0002-5208-9900} \and %
Maciej Wołczyk\inst{1}\orcidlink{0000-0002-3933-9971} \and %
Kamil Adamczewski\inst{1}\orcidlink{0000-0002-2917-4392} \and %
Tomasz Trzciński\inst{1,4,5}\orcidlink{0000-0002-1486-8906} \and %
Bartosz Zieliński\inst{1,2}\orcidlink{0000-0002-3063-3621}} %

\authorrunning{A.~Pardyl et al.}

\institute{IDEAS NCBR \\
\email{\{adam.pardyl, maciej.wolczyk, kamil.adamczewski,\\tomasz.trzcinski, bartosz.zielinski\}@ideas-ncbr.pl} \and
Jagiellonian University, Faculty of Mathematics and Computer Science \\
\email{michal.wronka@student.uj.edu.pl}\and
Jagiellonian University, Doctoral School of Exact and Natural Sciences \and
Warsaw University of Technology \and
Tooploox}

\maketitle

In this Supplementary Materials, we present additional experiments and visualizations for our \ours{} active visual exploration method.

\begin{table}[th]
\caption{\textbf{Heuristic baselines:} We compared our reinforcement learning-based method against two heuristic exploration policies for the reconstruction task on the ImageNet-1k dataset. Both heuristic methods use the same ElasticViT backbone as our approach and utilize $32^2$ glimpses, consisting of four ViT patches. The first method, called BasicElasticAME, samples one low-resolution glimpse of the entire image, followed by small, high-resolution glimpses selected using the AME algorithm. The second method, called QuadTreeAME, also begins with a single low-resolution glimpse. It then iteratively selects the most uncertain patch using AME, sampling a new glimpse at its position to effectively acquire four new higher-resolution patches in place of the selected lower-resolution patch. For comparison, we also provide results from a standard, fixed-resolution AME baseline. We observed that our method outperforms these heuristic baselines.}
\vspace{-.2cm}
\label{tab:segmentation}
\begin{center}
\begin{small}
\begin{tabular}{l@{\hskip6pt}llr@{\hskip8pt}cr}
\toprule
Method & RMSE & Image res. & Glimpses & Regime & Pixel \% \\
\midrule
AME & 30.3 & $128 \times 256$ & $8 \times 32^2$ &simple & 25.00 \\
BasicElasticAME & 25.95 & $224 \times 224$ & $12 \times 32^2$ & adaptive (heuristic) & 24.49  \\
QuadTreeAME & 23.2 & $224 \times 224$ & $12 \times 32^2$ & adaptive (heuristic) & 24.49  \\
\textbf{Ours} & \textbf{14.5} & $224 \times 224$ & $12 \times 32^2$ & adaptive & 24.49 \\
\bottomrule
\end{tabular}
\end{small}
\end{center}
\end{table}

\begin{table}[th]
\caption{\textbf{Model and training configuration}: In this table we specify the model and training configuration details.}
\vspace{-.2cm}
\label{tab:segmentation}
\begin{center}
\begin{small}
\begin{tabular}{l@{\hskip2cm}r}
\toprule
Parameter name & Value \\
\midrule 
\multicolumn{2}{c}{Encoder} \\
\midrule 
ViT type & \texttt{base} \\
native patch size & $16$ \\
transformer embed dim & $768$ \\
transformer blocks & $12$ \\
attention heads & $12$ \\
mlp ratio & $4$ \\
\midrule 
\multicolumn{2}{c}{Decoder} \\
\midrule 
transformer embed dim & $512$ \\
transformer blocks & $8$ \\
attention heads & $16$ \\
mlp ratio & $4$ \\

\midrule 
\multicolumn{2}{c}{RL agent} \\
\midrule 
hidden dim & $256$ \\
action distribution & TanhNormal \\
sac target entropy value & $-3$ \\
sac initial alpha value & $1$ \\
\midrule 
\multicolumn{2}{c}{Training} \\
\midrule 
training epochs & 100 \\
backbone pre-training epochs & 600 \\
backbone lr & $10^{-5}$ \\
rl agent lr & $5 * 10^{-4}$ \\
backbone weight decay rate & $10^{-4}$ \\
rl agent weight decay rate & $10^{-2}$ \\
lr scheduler type & one cycle \\
lr warmup epochs & 10 \\
minimum lr & $10^{-8}$\\
initial random action batches & $10000$ \\
initial frozen backbone epochs & $10$ \\
rl loss function & L2 \\
rl batch size & $256$ \\
backbone batch size & $128$ \\
replay buffer size & $10000$ \\

\bottomrule
\end{tabular}
\end{small}
\end{center}
\end{table}

\begin{figure}[t]
\begin{center}
\begin{small}
\begin{tabular}{r@{\hskip2pt}c@{\hskip2pt}c@{\hskip2pt}c@{\hskip2pt}c@{\hskip2pt}c@{\hskip2pt}c@{\hskip2pt}c}
A) & \tabfigure{width=\smallerimg}{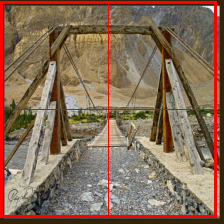} & \tabfigure{width=\smallerimg}{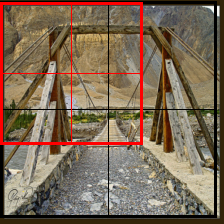} & \tabfigure{width=\smallerimg}{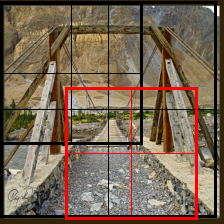} & \tabfigure{width=\smallerimg}{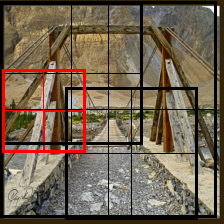} & \tabfigure{width=\smallerimg}{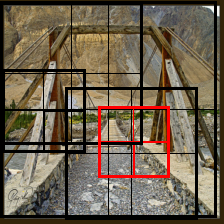} & \tabfigure{width=\smallerimg}{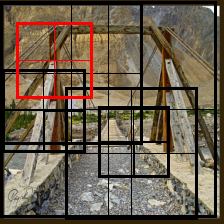} & \tabfigure{width=\smallerimg}{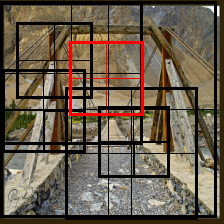}\\[19pt]
B) & \tabfigure{width=\smallerimg}{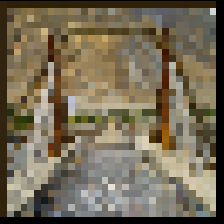} & \tabfigure{width=\smallerimg}{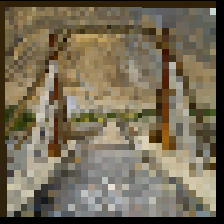} & \tabfigure{width=\smallerimg}{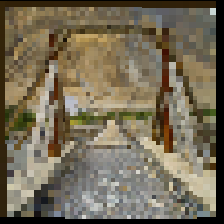} & \tabfigure{width=\smallerimg}{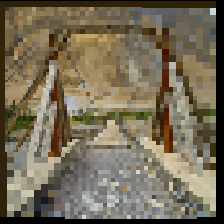} & \tabfigure{width=\smallerimg}{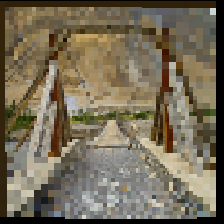} & \tabfigure{width=\smallerimg}{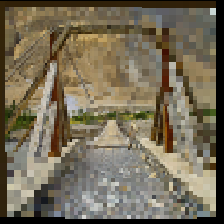} & \tabfigure{width=\smallerimg}{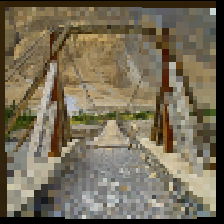}\\[19pt]
C) & brassiere & vault & gong & handrail & handrail & handrail & {\color{forestgreen4416044}bridge}  \\[2pt]
D) & 26\% & 11\% & 23\% & 9\% & 54\% & 47\% & {\color{forestgreen4416044}86\%} \\[15pt]
\end{tabular}
\begin{tabular}{r@{\hskip2pt}c@{\hskip2pt}c@{\hskip2pt}c@{\hskip2pt}c@{\hskip2pt}c}
A) & \tabfigure{width=\smallerimg}{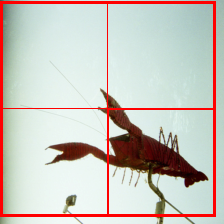} & \tabfigure{width=\smallerimg}{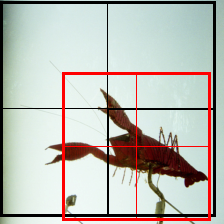} & \tabfigure{width=\smallerimg}{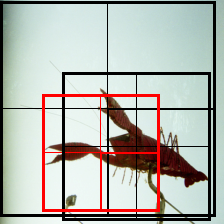} & \tabfigure{width=\smallerimg}{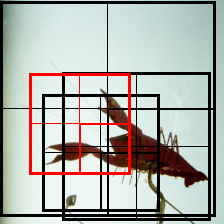} & \tabfigure{width=\smallerimg}{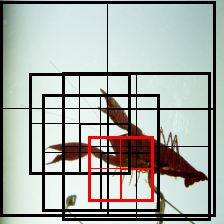} \\[19pt]
B) & \tabfigure{width=\smallerimg}{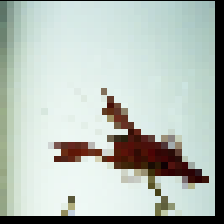} & \tabfigure{width=\smallerimg}{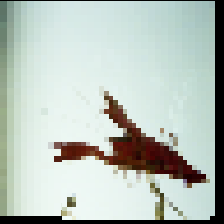} & \tabfigure{width=\smallerimg}{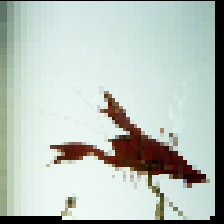} & \tabfigure{width=\smallerimg}{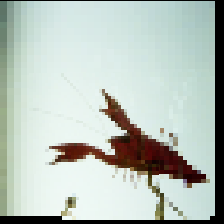} & \tabfigure{width=\smallerimg}{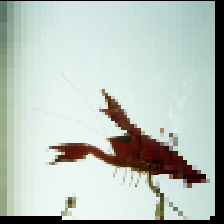} \\[19pt]
C) & dragonfly & {\color{forestgreen4416044}crayfish} & {\color{forestgreen4416044}crayfish} & {\color{forestgreen4416044}crayfish} & {\color{forestgreen4416044}crayfish} \\[2pt]
D) & 18\% & 17\% & 56\% & 78\% & {\color{forestgreen4416044}84\%} \\[15pt]
\end{tabular}
\begin{tabular}{r@{\hskip2pt}c@{\hskip2pt}c@{\hskip2pt}c@{\hskip2pt}c@{\hskip2pt}c@{\hskip2pt}c@{\hskip2pt}c}
A) & \tabfigure{width=\smallerimg}{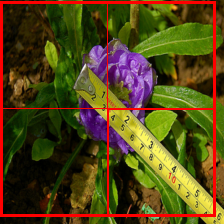} & \tabfigure{width=\smallerimg}{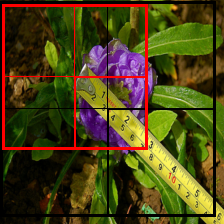} & \tabfigure{width=\smallerimg}{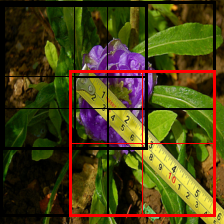} & \tabfigure{width=\smallerimg}{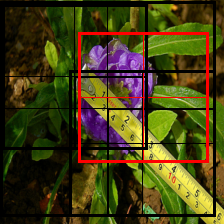} & \tabfigure{width=\smallerimg}{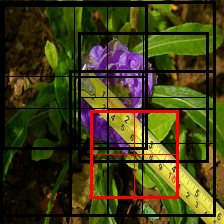} & \tabfigure{width=\smallerimg}{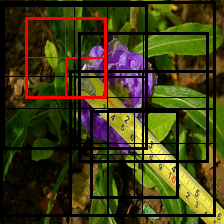} & \tabfigure{width=\smallerimg}{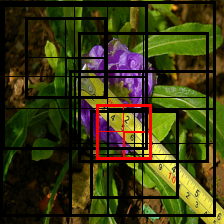}\\[19pt]
B) & \tabfigure{width=\smallerimg}{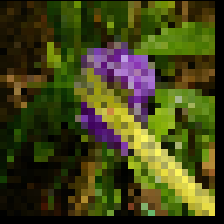} & \tabfigure{width=\smallerimg}{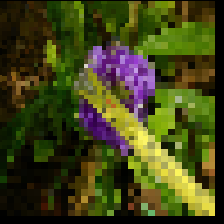} & \tabfigure{width=\smallerimg}{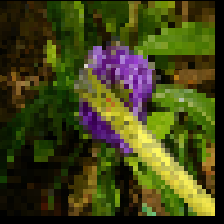} & \tabfigure{width=\smallerimg}{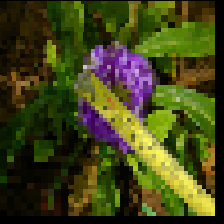} & \tabfigure{width=\smallerimg}{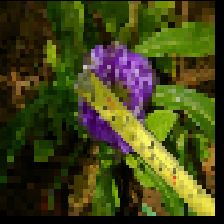} & \tabfigure{width=\smallerimg}{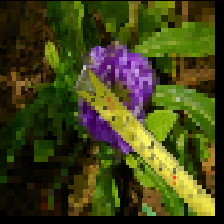} & \tabfigure{width=\smallerimg}{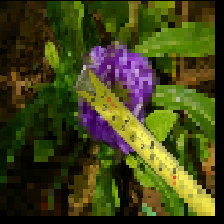}\\[19pt]
C) & \makecell[c]{cabbage\\butterfly} & \makecell[c]{cabbage\\butterfly} & dragonfly & {\color{forestgreen4416044}ruler} & {\color{forestgreen4416044}ruler} & {\color{forestgreen4416044}ruler} & {\color{forestgreen4416044}ruler}  \\[6pt]
D) & 10\% & 27\% & 29\% & 49\% & 71\% & 76\% & {\color{forestgreen4416044}85\%} \\[15pt]
\end{tabular}
\end{small}
\end{center}
\vspace{-.2cm}
\caption{\textbf{Image classification on ImageNet-1k step-by-step:} \ours{} explores $224 \times 224$ images from ImageNet with $32 \times 32$ glimpses of variable scale, zooming in on objects of interest and stopping the process after reaching $75\%$ predicted probability. The rows correspond to: A) glimpse locations, B) pixels visible to the model (interpolated from glimpses for preview), C) predicted label, D) prediction probability.}
\label{fig:qualiative-cls1}
\end{figure}
\begin{figure}[t]
\begin{center}
\begin{small}
\begin{tabular}{r@{\hskip2pt}c@{\hskip2pt}c@{\hskip2pt}c}
A) & \tabfigure{width=\smallerimg}{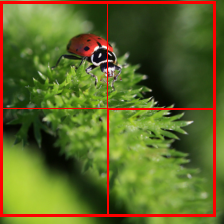} & \tabfigure{width=\smallerimg}{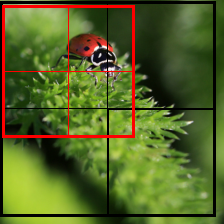} & \tabfigure{width=\smallerimg}{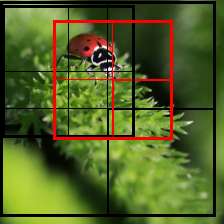} \\[19pt]
B) & \tabfigure{width=\smallerimg}{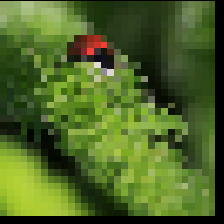} & \tabfigure{width=\smallerimg}{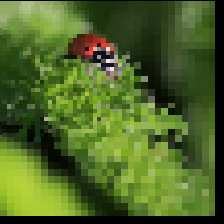} & \tabfigure{width=\smallerimg}{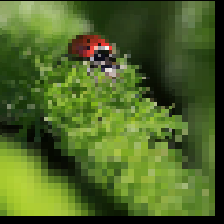} \\[19pt]
C) & {\color{forestgreen4416044}ladybug}  & {\color{forestgreen4416044}ladybug}  & {\color{forestgreen4416044}ladybug}  \\[2pt]
D) & 62\% & 77\% & {\color{forestgreen4416044}80\%} \\[15pt]
\end{tabular}
\begin{tabular}{r@{\hskip2pt}c@{\hskip2pt}c@{\hskip2pt}c@{\hskip2pt}c@{\hskip2pt}c@{\hskip2pt}c@{\hskip2pt}c}
A) & \tabfigure{width=\smallerimg}{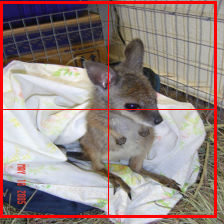} & \tabfigure{width=\smallerimg}{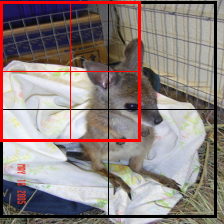} & \tabfigure{width=\smallerimg}{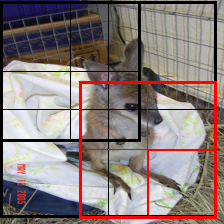} & \tabfigure{width=\smallerimg}{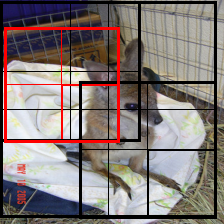} & \tabfigure{width=\smallerimg}{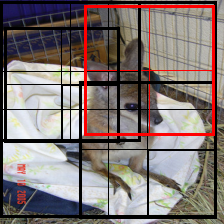} & \tabfigure{width=\smallerimg}{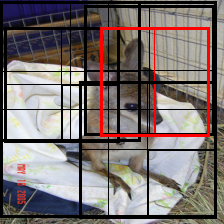} & \tabfigure{width=\smallerimg}{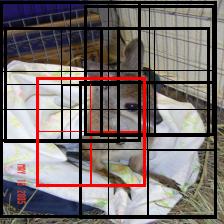}\\[19pt]
B) & \tabfigure{width=\smallerimg}{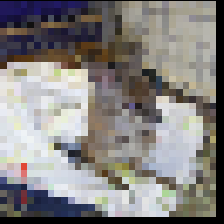} & \tabfigure{width=\smallerimg}{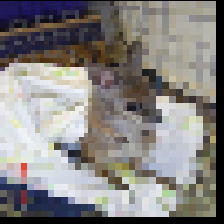} & \tabfigure{width=\smallerimg}{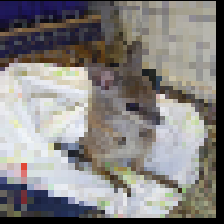} & \tabfigure{width=\smallerimg}{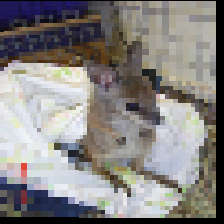} & \tabfigure{width=\smallerimg}{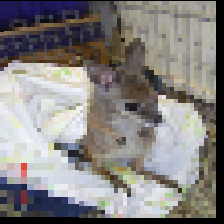} & \tabfigure{width=\smallerimg}{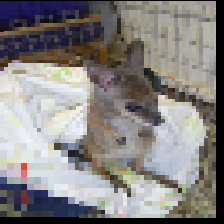} & \tabfigure{width=\smallerimg}{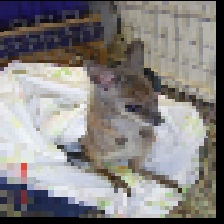}\\[19pt]
C) & hamper & napkin & hamster & \makecell[c]{wood\\rabbit} & \makecell[c]{wood\\rabbit} & \makecell[c]{wood\\rabbit} & \makecell[c]{wood\\rabbit} \\[6pt]
& & & & & & {\color{crimson2143940}should be:}&{\color{crimson2143940}wallaby} \\
D) & 8\% & 7\% & 13\% & 52\% & 78\% & 79\% & {\color{forestgreen4416044}85\%} \\[15pt]
\end{tabular}
\begin{tabular}{r@{\hskip2pt}c@{\hskip2pt}c@{\hskip2pt}c@{\hskip2pt}c}
A) & \tabfigure{width=\smallerimg}{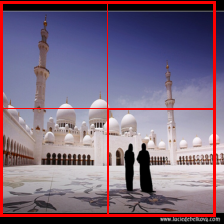} & \tabfigure{width=\smallerimg}{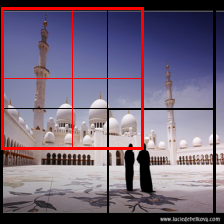} & \tabfigure{width=\smallerimg}{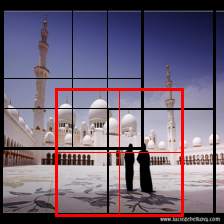} & \tabfigure{width=\smallerimg}{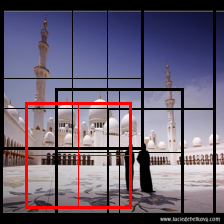} \\[19pt]
B) & \tabfigure{width=\smallerimg}{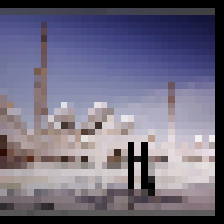} & \tabfigure{width=\smallerimg}{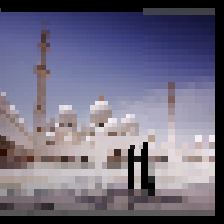} & \tabfigure{width=\smallerimg}{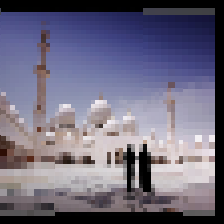} & \tabfigure{width=\smallerimg}{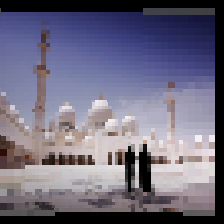} \\[19pt]
C) & {\color{forestgreen4416044}mosque} & fountain & {\color{forestgreen4416044}mosque} & {\color{forestgreen4416044}mosque} \\[6pt]
D) & 16\% & 46\% & 21\% & {\color{forestgreen4416044}90\%} \\[15pt]
\end{tabular}
\end{small}
\end{center}
\vspace{-.2cm}
\caption{\textbf{Image classification on ImageNet-1k step-by-step:} \ours{} explores $224 \times 224$ images from ImageNet with $32 \times 32$ glimpses of variable scale, zooming in on objects of interest and stopping the process after reaching $75\%$ predicted probability. The rows correspond to: A) glimpse locations, B) pixels visible to the model (interpolated from glimpses for preview), C) predicted label, D) prediction probability.}
\label{fig:qualiative-cls2}
\end{figure}

\begin{figure}[t]
\begin{center}
\begin{small}
\begin{tabular}{rc@{\hskip2pt}c@{\hskip2pt}c@{\hskip2pt}|@{\hskip2pt}rc@{\hskip2pt}c@{\hskip2pt}c}
& A) Location & B) Visible & C) Output && A) Location & B) Visible & C) Output \\
1 & \tabfigure{width=\smalllimg}{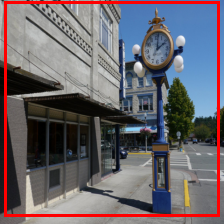} & \tabfigure{width=\smalllimg}{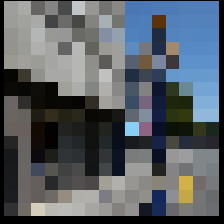} & \tabfigure{width=\smalllimg}{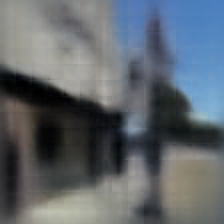} & 7 & \tabfigure{width=\smalllimg}{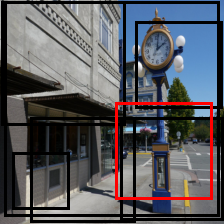} & \tabfigure{width=\smalllimg}{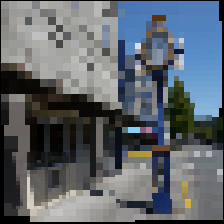} & \tabfigure{width=\smalllimg}{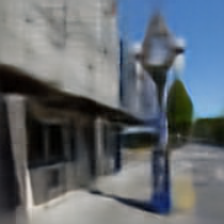} \\[25pt]
2 & \tabfigure{width=\smalllimg}{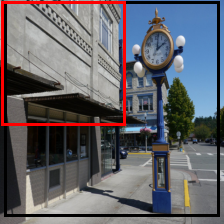} & \tabfigure{width=\smalllimg}{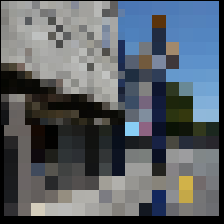} & \tabfigure{width=\smalllimg}{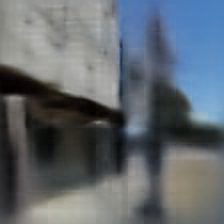} & 8 & \tabfigure{width=\smalllimg}{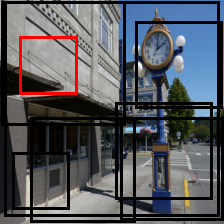} & \tabfigure{width=\smalllimg}{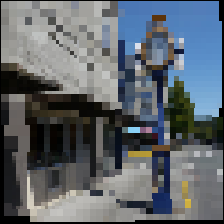} & \tabfigure{width=\smalllimg}{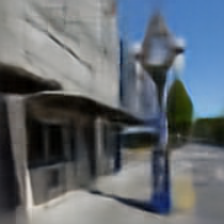} \\[25pt]
3&\tabfigure{width=\smalllimg}{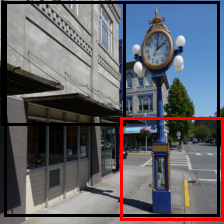} & \tabfigure{width=\smalllimg}{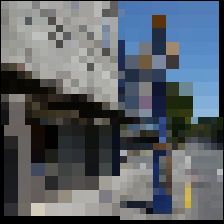} & \tabfigure{width=\smalllimg}{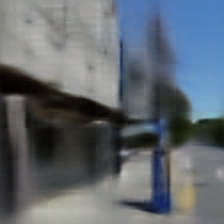} & 9 & \tabfigure{width=\smalllimg}{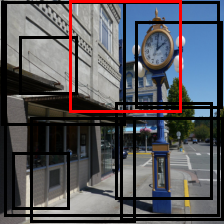} & \tabfigure{width=\smalllimg}{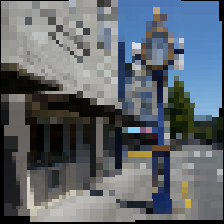} & \tabfigure{width=\smalllimg}{} \\[25pt]
4&\tabfigure{width=\smalllimg}{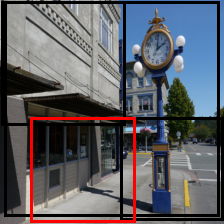} & \tabfigure{width=\smalllimg}{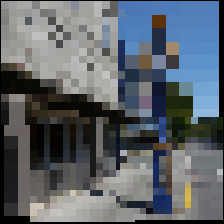} & \tabfigure{width=\smalllimg}{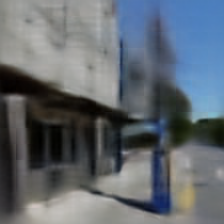} & 10 & \tabfigure{width=\smalllimg}{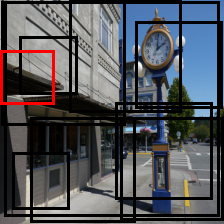} & \tabfigure{width=\smalllimg}{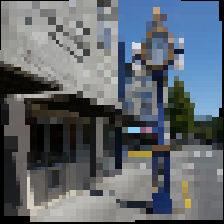} & \tabfigure{width=\smalllimg}{} \\[25pt]
5&\tabfigure{width=\smalllimg}{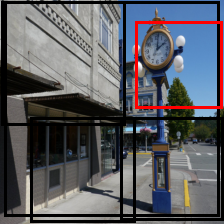} & \tabfigure{width=\smalllimg}{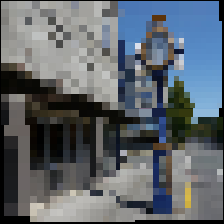} & \tabfigure{width=\smalllimg}{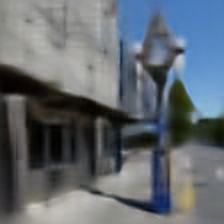} & 11 & \tabfigure{width=\smalllimg}{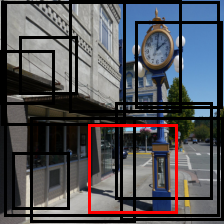} & \tabfigure{width=\smalllimg}{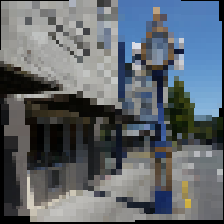} & \tabfigure{width=\smalllimg}{} \\[25pt]
6&\tabfigure{width=\smalllimg}{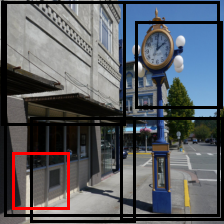} & \tabfigure{width=\smalllimg}{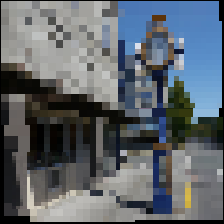} & \tabfigure{width=\smalllimg}{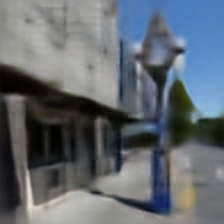} & 12 & \tabfigure{width=\smalllimg}{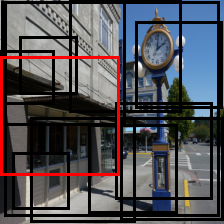} & \tabfigure{width=\smalllimg}{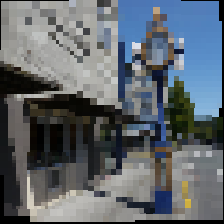} & \tabfigure{width=\smalllimg}{} \\[25pt]
\end{tabular}
\end{small}
\end{center}
\vspace{-.2cm}
\caption{\textbf{Image reconstruction on MS COCO step-by-step:} \ours{} explores $224 \times 224$ images from MS COCO with $8\times16^2$ glimpses of variable scale, zooming in on objects of interest. Note, that each glimpse consists of a single vision transformer patch. The columns correspond to: A) glimpse locations, B) pixels visible to the model (interpolated from glimpses for preview), C) reconstruction result.}
\label{fig:qualiative-rec1}
\end{figure}

\begin{figure}[t]
\begin{center}
\begin{small}
\begin{tabular}{rc@{\hskip2pt}c@{\hskip2pt}c@{\hskip2pt}|@{\hskip2pt}rc@{\hskip2pt}c@{\hskip2pt}c}
& A) Location & B) Visible & C) Output && A) Location & B) Visible & C) Output \\
1 & \tabfigure{width=\smalllimg}{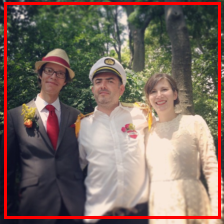} & \tabfigure{width=\smalllimg}{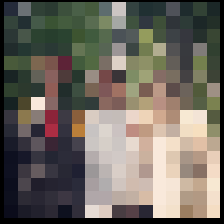} & \tabfigure{width=\smalllimg}{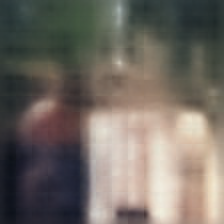} & 7 & \tabfigure{width=\smalllimg}{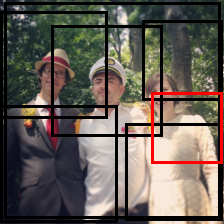} & \tabfigure{width=\smalllimg}{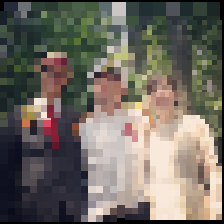} & \tabfigure{width=\smalllimg}{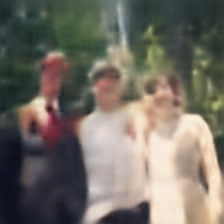} \\[25pt]
2 & \tabfigure{width=\smalllimg}{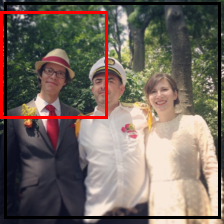} & \tabfigure{width=\smalllimg}{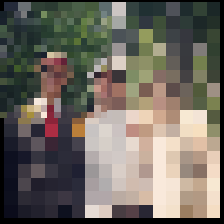} & \tabfigure{width=\smalllimg}{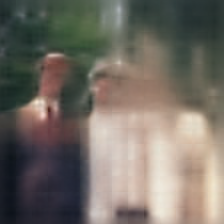} & 8 & \tabfigure{width=\smalllimg}{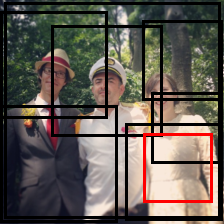} & \tabfigure{width=\smalllimg}{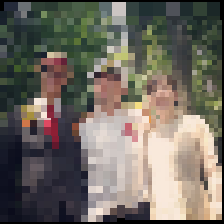} & \tabfigure{width=\smalllimg}{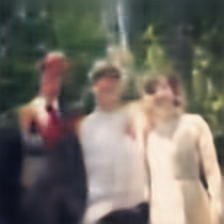} \\[25pt]
3&\tabfigure{width=\smalllimg}{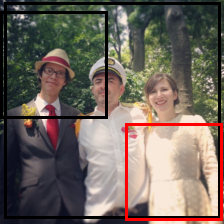} & \tabfigure{width=\smalllimg}{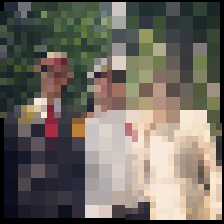} & \tabfigure{width=\smalllimg}{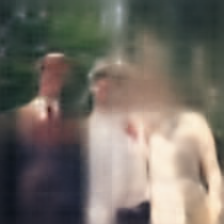} & 9 & \tabfigure{width=\smalllimg}{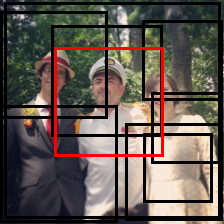} & \tabfigure{width=\smalllimg}{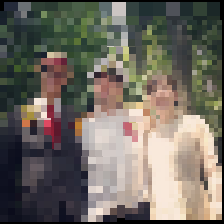} & \tabfigure{width=\smalllimg}{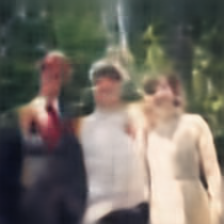} \\[25pt]
4&\tabfigure{width=\smalllimg}{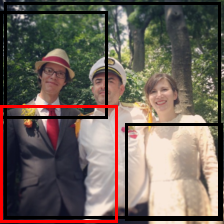} & \tabfigure{width=\smalllimg}{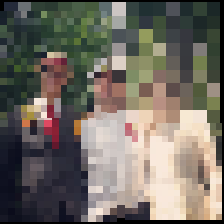} & \tabfigure{width=\smalllimg}{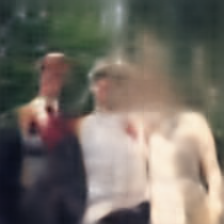} & 10 & \tabfigure{width=\smalllimg}{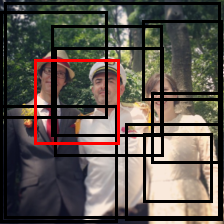} & \tabfigure{width=\smalllimg}{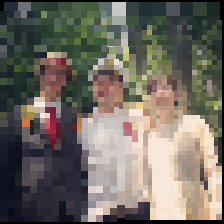} & \tabfigure{width=\smalllimg}{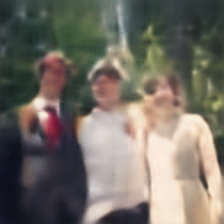} \\[25pt]
5&\tabfigure{width=\smalllimg}{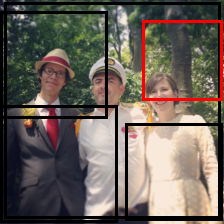} & \tabfigure{width=\smalllimg}{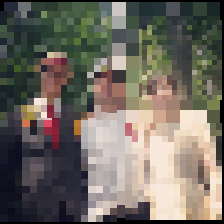} & \tabfigure{width=\smalllimg}{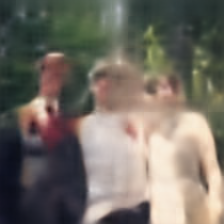} & 11 & \tabfigure{width=\smalllimg}{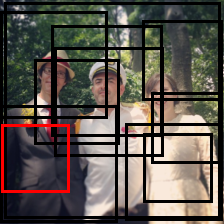} & \tabfigure{width=\smalllimg}{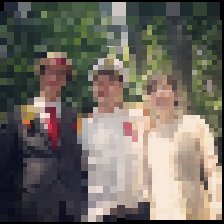} & \tabfigure{width=\smalllimg}{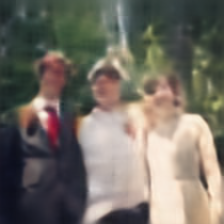} \\[25pt]
6&\tabfigure{width=\smalllimg}{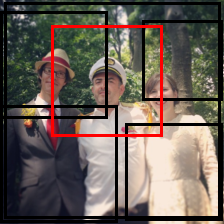} & \tabfigure{width=\smalllimg}{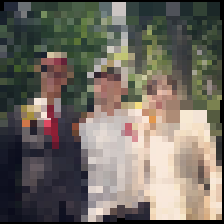} & \tabfigure{width=\smalllimg}{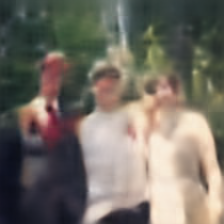} & 12 & \tabfigure{width=\smalllimg}{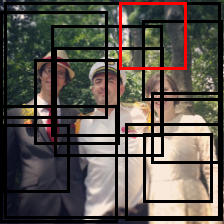} & \tabfigure{width=\smalllimg}{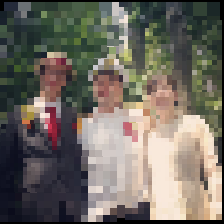} & \tabfigure{width=\smalllimg}{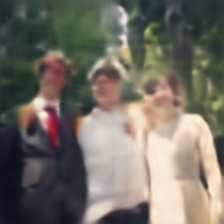} \\[25pt]
\end{tabular}
\end{small}
\end{center}
\vspace{-.2cm}
\caption{\textbf{Image reconstruction on MS COCO step-by-step:} \ours{} explores $224 \times 224$ images from MS COCO with $8\times16^2$ glimpses of variable scale, zooming in on objects of interest. Note, that each glimpse consists of a single vision transformer patch. The columns correspond to: A) glimpse locations, B) pixels visible to the model (interpolated from glimpses for preview), C) reconstruction result.}
\label{fig:qualiative-rec2}
\end{figure}

\begin{figure}[t]
\begin{center}
\begin{small}
\begin{tabular}{rc@{\hskip2pt}c@{\hskip2pt}c@{\hskip2pt}|@{\hskip2pt}rc@{\hskip2pt}c@{\hskip2pt}c}
& A) Location & B) Visible & C) Output && A) Location & B) Visible & C) Output \\
1 & \tabfigure{width=\smalllimg}{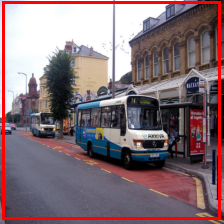} & \tabfigure{width=\smalllimg}{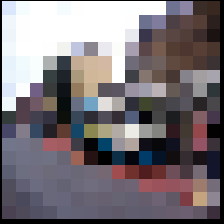} & \tabfigure{width=\smalllimg}{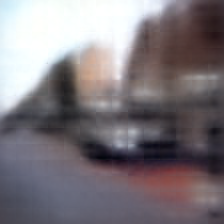} & 7 & \tabfigure{width=\smalllimg}{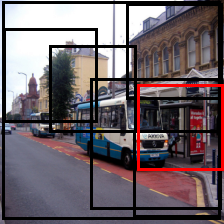} & \tabfigure{width=\smalllimg}{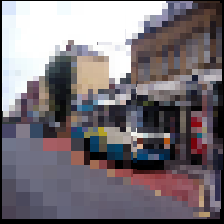} & \tabfigure{width=\smalllimg}{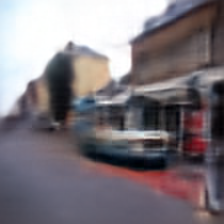} \\[25pt]
2 & \tabfigure{width=\smalllimg}{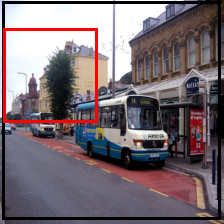} & \tabfigure{width=\smalllimg}{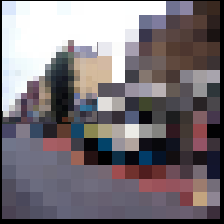} & \tabfigure{width=\smalllimg}{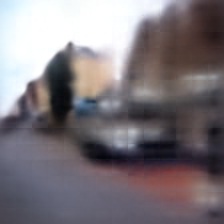} & 8 & \tabfigure{width=\smalllimg}{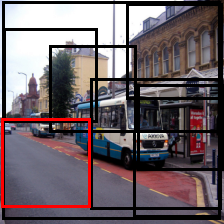} & \tabfigure{width=\smalllimg}{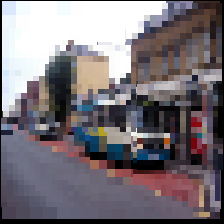} & \tabfigure{width=\smalllimg}{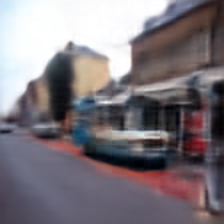} \\[25pt]
3&\tabfigure{width=\smalllimg}{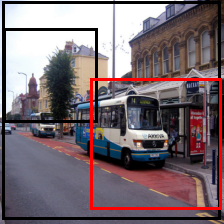} & \tabfigure{width=\smalllimg}{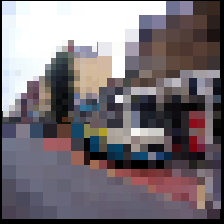} & \tabfigure{width=\smalllimg}{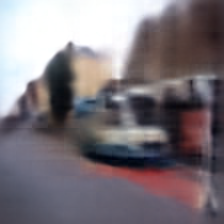} & 9 & \tabfigure{width=\smalllimg}{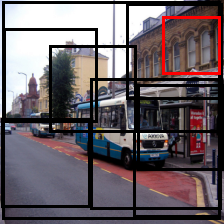} & \tabfigure{width=\smalllimg}{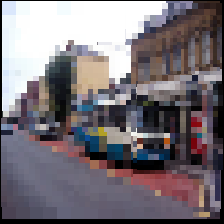} & \tabfigure{width=\smalllimg}{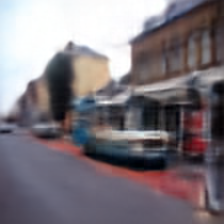} \\[25pt]
4&\tabfigure{width=\smalllimg}{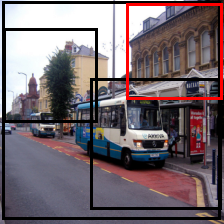} & \tabfigure{width=\smalllimg}{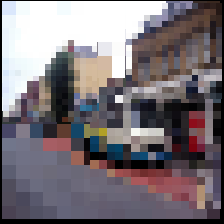} & \tabfigure{width=\smalllimg}{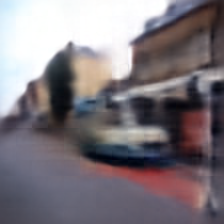} & 10 & \tabfigure{width=\smalllimg}{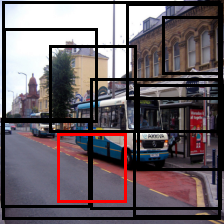} & \tabfigure{width=\smalllimg}{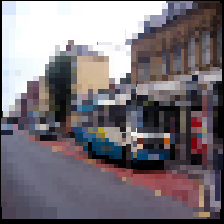} & \tabfigure{width=\smalllimg}{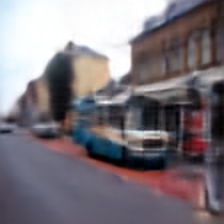} \\[25pt]
5&\tabfigure{width=\smalllimg}{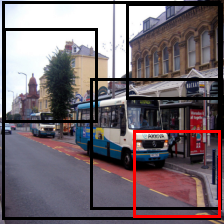} & \tabfigure{width=\smalllimg}{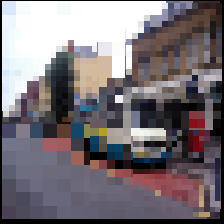} & \tabfigure{width=\smalllimg}{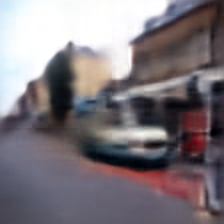} & 11 & \tabfigure{width=\smalllimg}{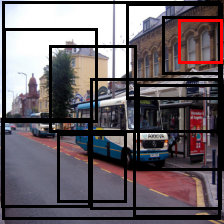} & \tabfigure{width=\smalllimg}{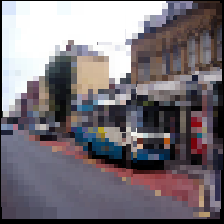} & \tabfigure{width=\smalllimg}{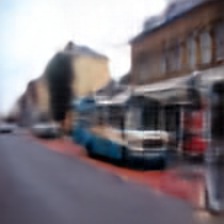} \\[25pt]
6&\tabfigure{width=\smalllimg}{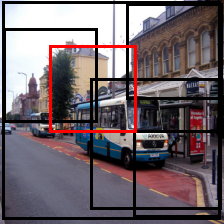} & \tabfigure{width=\smalllimg}{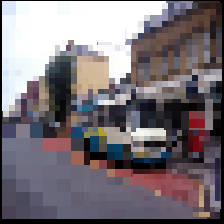} & \tabfigure{width=\smalllimg}{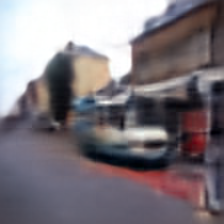} & 12 & \tabfigure{width=\smalllimg}{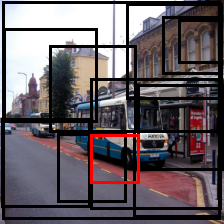} & \tabfigure{width=\smalllimg}{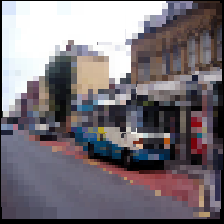} & \tabfigure{width=\smalllimg}{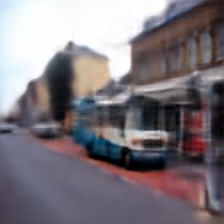} \\[25pt]
\end{tabular}
\end{small}
\end{center}
\vspace{-.2cm}
\caption{\textbf{Image reconstruction on MS COCO step-by-step:} \ours{} explores $224 \times 224$ images from MS COCO with $8\times16^2$ glimpses of variable scale, zooming in on objects of interest. Note, that each glimpse consists of a single vision transformer patch. The columns correspond to: A) glimpse locations, B) pixels visible to the model (interpolated from glimpses for preview), C) reconstruction result.}
\label{fig:qualiative-rec3}
\end{figure}

\begin{figure}[t]
\begin{center}
\begin{small}
\begin{sc}
\begin{tabular}{cc@{\hskip10pt}|@{\hskip10pt}cc}
Image & Reconstruction & Image & Reconstruction \\
\includegraphics[width=\medimg]{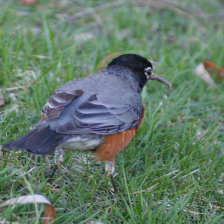} & \includegraphics[width=\medimg]{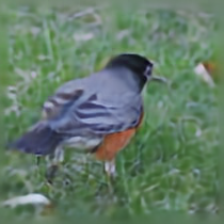} & \includegraphics[width=\medimg]{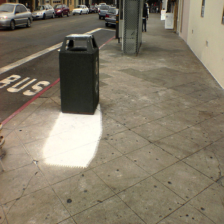} & \includegraphics[width=\medimg]{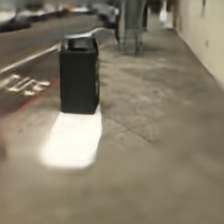} \\[4pt]
\includegraphics[width=\medimg]{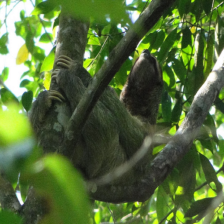} & \includegraphics[width=\medimg]{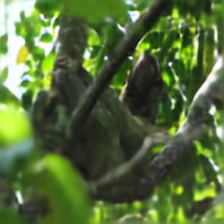} & \includegraphics[width=\medimg]{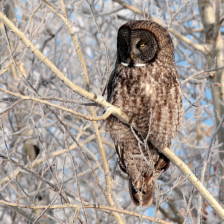} & \includegraphics[width=\medimg]{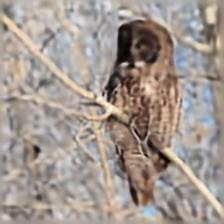} \\[4pt]
\includegraphics[width=\medimg]{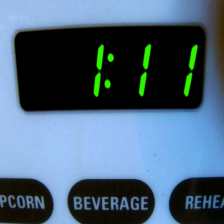} & \includegraphics[width=\medimg]{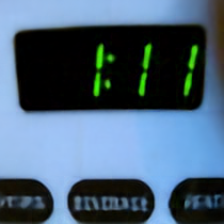} & \includegraphics[width=\medimg]{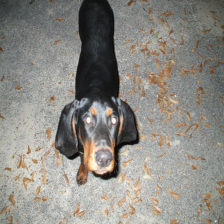} & \includegraphics[width=\medimg]{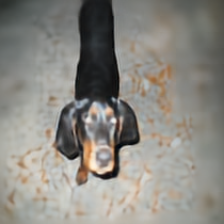} \\[4pt]
\includegraphics[width=\medimg]{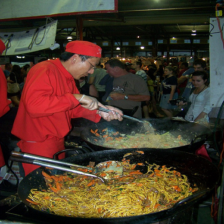} & \includegraphics[width=\medimg]{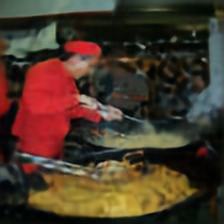} & \includegraphics[width=\medimg]{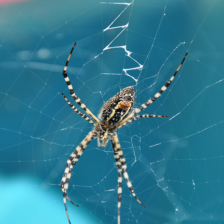} & \includegraphics[width=\medimg]{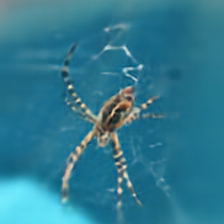} \\[4pt]
\includegraphics[width=\medimg]{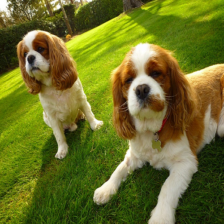} & \includegraphics[width=\medimg]{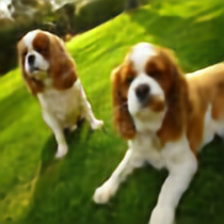} & \includegraphics[width=\medimg]{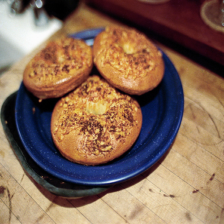} & \includegraphics[width=\medimg]{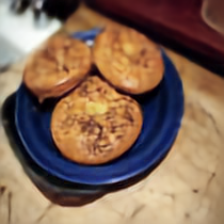} \\
\end{tabular}
\end{sc}
\end{small}
\end{center}
\vskip -0.1in
\caption{\textbf{Image reconstruction on ImageNet-1k:} Figure shows the qualitative results of the image reconstruction task. Image size is $224\times224$ and $12\times32^2$ adaptive glimpses are used.}
\end{figure}

\begin{figure}[t]
\begin{center}
\begin{small}
\begin{sc}
\begin{tabular}{cc@{\hskip10pt}|@{\hskip10pt}cc}
Image & Reconstruction & Image & Reconstruction \\
\includegraphics[width=\medimg]{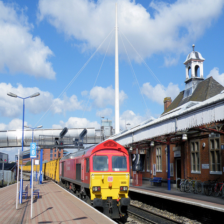} & \includegraphics[width=\medimg]{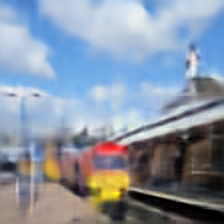} & \includegraphics[width=\medimg]{} & \includegraphics[width=\medimg]{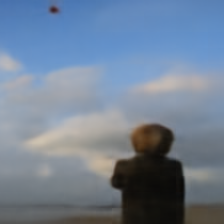} \\[4pt]
\includegraphics[width=\medimg]{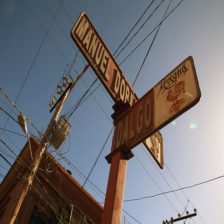} & \includegraphics[width=\medimg]{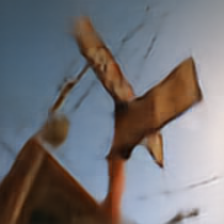} & \includegraphics[width=\medimg]{} & \includegraphics[width=\medimg]{} \\[4pt]
\includegraphics[width=\medimg]{} & \includegraphics[width=\medimg]{} & \includegraphics[width=\medimg]{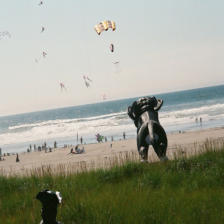} & \includegraphics[width=\medimg]{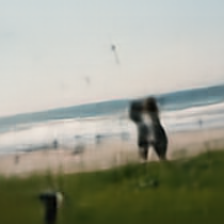} \\[4pt]
\includegraphics[width=\medimg]{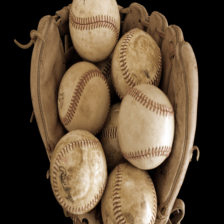} & \includegraphics[width=\medimg]{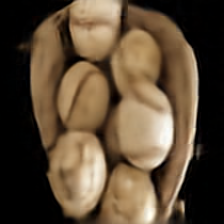} & \includegraphics[width=\medimg]{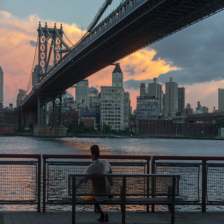} & \includegraphics[width=\medimg]{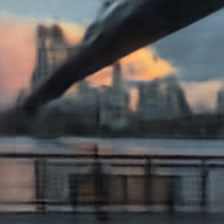} \\[4pt]
\includegraphics[width=\medimg]{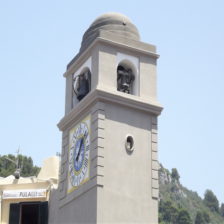} & \includegraphics[width=\medimg]{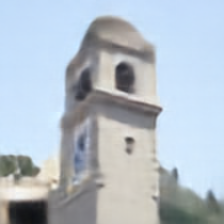} & \includegraphics[width=\medimg]{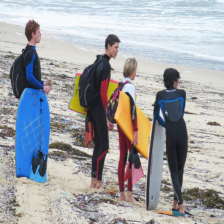} & \includegraphics[width=\medimg]{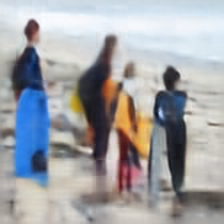} \\
\end{tabular}
\end{sc}
\end{small}
\end{center}
\vskip -0.1in
\caption{\textbf{Image reconstruction on MS COCO:} Figure shows the qualitative results of the image reconstruction task. Image size is $224\times224$ and $12\times16^2$ adaptive glimpses are used.}
\end{figure}

\begin{figure}[t]
\begin{center}
\begin{small}
\begin{sc}
\begin{tabular}{cc}
Image & Reconstruction \\
\includegraphics[width=\smallimgT,height=\smallimg]{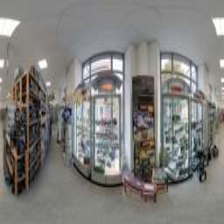} & \includegraphics[width=\smallimgT,height=\smallimg]{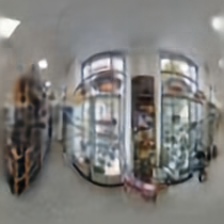} \\[4pt]\includegraphics[width=\smallimgT,height=\smallimg]{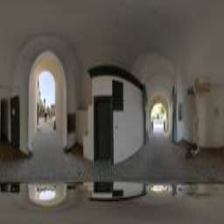} & \includegraphics[width=\smallimgT,height=\smallimg]{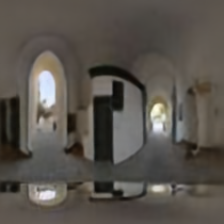} \\
\includegraphics[width=\smallimgT,height=\smallimg]{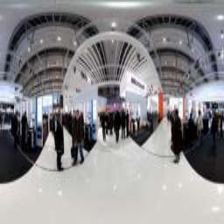} & \includegraphics[width=\smallimgT,height=\smallimg]{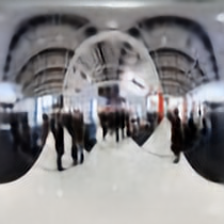} \\[4pt]\includegraphics[width=\smallimgT,height=\smallimg]{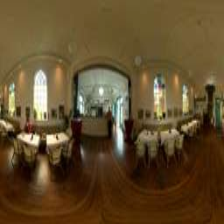} & \includegraphics[width=\smallimgT,height=\smallimg]{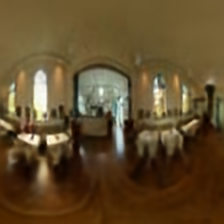} \\
\includegraphics[width=\smallimgT,height=\smallimg]{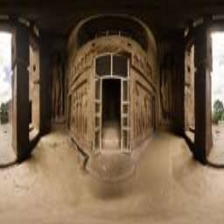} & \includegraphics[width=\smallimgT,height=\smallimg]{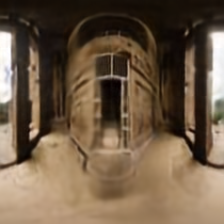} \\[4pt]\includegraphics[width=\smallimgT,height=\smallimg]{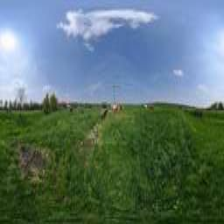} & \includegraphics[width=\smallimgT,height=\smallimg]{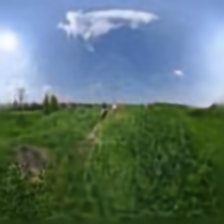} \\
\end{tabular}
\end{sc}
\end{small}
\end{center}
\vskip -0.1in
\caption{\textbf{Image reconstruction on SUN360:} Figure shows the qualitative results of the image reconstruction task. Image size is $224\times224$ and $12\times32^2$ adaptive glimpses are used. For visualization, images were resized to the original aspect ratio.}
\end{figure}

\begin{figure}[t]
\begin{center}
\begin{small}
\begin{sc}
\begin{tabular}{ccc}
Image & Target & Result \\
\includegraphics[width=\medimg]{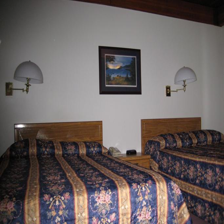} & \includegraphics[width=\medimg]{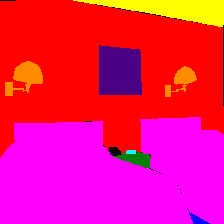} & \includegraphics[width=\medimg]{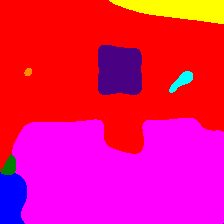} \\
\includegraphics[width=\medimg]{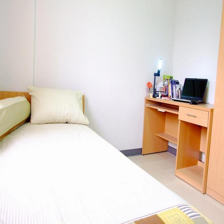} & \includegraphics[width=\medimg]{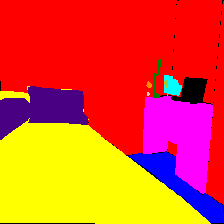} & \includegraphics[width=\medimg]{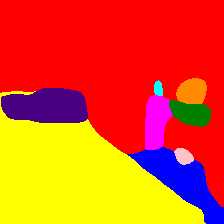} \\
\includegraphics[width=\medimg]{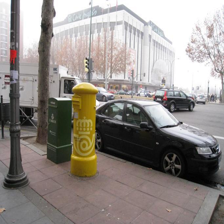} & \includegraphics[width=\medimg]{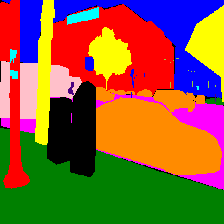} & \includegraphics[width=\medimg]{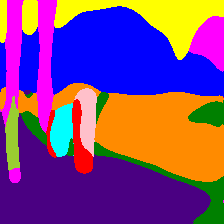} \\
\includegraphics[width=\medimg]{} & \includegraphics[width=\medimg]{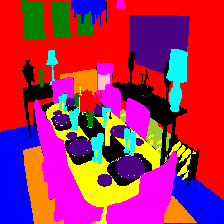} & \includegraphics[width=\medimg]{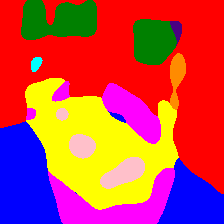} \\
\includegraphics[width=\medimg]{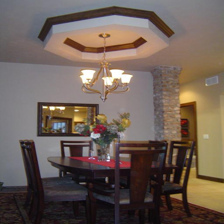} & \includegraphics[width=\medimg]{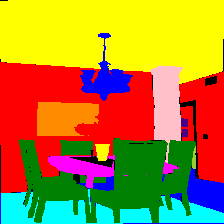} & \includegraphics[width=\medimg]{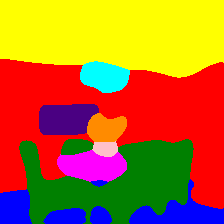} \\
\includegraphics[width=\medimg]{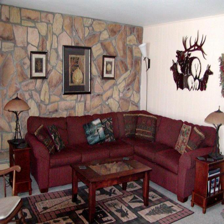} & \includegraphics[width=\medimg]{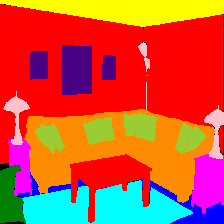} & \includegraphics[width=\medimg]{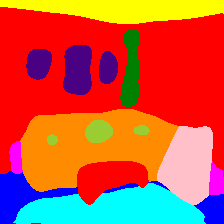}
\end{tabular}
\end{sc}
\end{small}
\end{center}
\vskip -0.1in
\caption{\textbf{Image segmentation on ADE20K:} Figure shows the qualitative results of the image segmentation task. Image size is $224\times224$ and $12\times48^2$ adaptive glimpses are used.}
\end{figure}

\clearpage

%
%